%% file: main.tex
\documentclass{article}

     \PassOptionsToPackage{numbers,sort&compress}{natbib}

\usepackage[final]{neurips_2025}

\usepackage[utf8]{inputenc} 
\usepackage[T1]{fontenc}    
\usepackage{url}            
\usepackage{booktabs}       
\usepackage{amsfonts}       
\usepackage{nicefrac}       
\usepackage{microtype}      
\usepackage{xcolor}         
\usepackage{verbatim}
\usepackage{graphicx}
\usepackage{amsmath}
\usepackage{amssymb}
\usepackage{tikz}
\usepackage{pgfplots}
\usepackage{multirow}
\usepackage{color}
\usepackage{array}
\usepackage{pifont}
\usepackage{times}
\usepackage{epsfig}
\usepackage{lipsum}
\usepackage{subfiles}
\usepackage{subcaption}
\pgfplotsset{compat=1.16}
\usepackage{enumitem}
\usepackage{float}
\usepackage{bm}
\usepackage{cuted}
\usepackage{wrapfig}
\usepackage{colortbl}

\usepackage{xspace}
\makeatletter
\DeclareRobustCommand\onedot{\futurelet\@let@token\@onedot}
\def\@onedot{\ifx\@let@token.\else.\null\fi\xspace}

\def\eg{\emph{e.g}\onedot}

 \def\vs{\emph{vs}\onedot}

\makeatother

\DeclareRobustCommand{\ours}{VFMTok\xspace}

\newlength\savewidth\newcommand\shline{\noalign{\global\savewidth\arrayrulewidth
  \global\arrayrulewidth 1pt}\hline\noalign{\global\arrayrulewidth\savewidth}}

\makeatletter
\newcommand{\smaller}{\@setfontsize\mynotesize{8.4pt}{9.4pt}}
\makeatother

\newcommand{\tablestyle}[2]{\setlength{\tabcolsep}{#1}\renewcommand{\arraystretch}{#2}\centering\small}

\newcommand{\tablestylesmaller}[2]{\setlength{\tabcolsep}{#1}\renewcommand{\arraystretch}{#2}\centering\smaller}

\definecolor{citeblue}{HTML}{5B9BD5}
\usepackage[colorlinks=true]{hyperref} 
\hypersetup{
    citecolor=citeblue,   
    linkcolor=red,   
    urlcolor=citeblue     
}

\usepackage[capitalize]{cleveref}
\crefname{section}{Sec.}{Secs.}
\Crefname{section}{Section}{Sections}
\Crefname{table}{Table}{Tables}
\crefname{table}{Tab.}{Tabs.}
\Crefname{appendix}{Appendix}{Appendices}
\crefname{appendix}{Appx.}{Appxs.}

\title{Vision Foundation Models as Effective Visual Tokenizers for Autoregressive Generation}

\author{
    \vspace{-8mm} \\
    \textbf{
    Anlin Zheng$^{1}$
    \quad Xin Wen$^{1}$
    \quad Xuanyang Zhang$^{2\ast}$
    \quad Chuofan Ma$^{1}$
    } \vspace{1mm} \\
    \textbf{
    Tiancai Wang$^{3}$
    \quad Gang Yu$^{2}$
    \quad Xiangyu Zhang$^{2,4}$
    \quad Xiaojuan Qi$^{1\ast\dagger}$ 
    } \vspace{2mm} \\
    $^1$The University of Hong Kong\quad $^2$StepFun \quad $^3$Dexmal \quad
    $^4$MEGVII Technology \vspace{-4mm}
}

\begin{document}

\maketitle
{\let\thefootnote\relax\footnotetext{\noindent $^{\dagger}$~Corresponding author: \texttt{xjqi@eee.hku.hk}. $^{\ast}$~Project lead.}}




\input{sections/0_abstract}
\input{sections/1_introduction}
\input{sections/2_related_work}

\input{sections/3_approach}
\input{sections/4_experiments}

\input{sections/5_conclusion}
{
    \small
    \bibliographystyle{plain}
    \bibliography{main}
}
\clearpage
\input{sections/6_appendix}

\end{document}

%% file: sections/0_abstract.tex
\begin{abstract}

In this work, we present a novel direction to build an image tokenizer directly on top of a frozen vision foundation model, which is a largely underexplored area. Specifically, we employ a frozen vision foundation model as the encoder of our tokenizer. To enhance its effectiveness, we introduce two key components: (1) a region-adaptive quantization framework that reduces redundancy in the pre-trained features on regular 2D grids, and (2) a semantic reconstruction objective that aligns the tokenizer’s outputs with the foundation model’s representations to preserve semantic fidelity. Based on these designs, our proposed image tokenizer, \textbf{\ours}, achieves substantial improvements in image reconstruction and generation quality, while also enhancing token efficiency. It further boosts autoregressive (AR) generation---achieving a gFID of \textbf{1.36} on ImageNet benchmarks, while accelerating model convergence by \textbf{three times}, and enabling high-fidelity class-conditional synthesis without the need for classifier-free guidance (CFG). The code is available at \href{https://github.com/CVMI-Lab/VFMTok}{https://github.com/CVMI-Lab/VFMTok}.

\end{abstract}

%% file: sections/1_introduction.tex
\section{Introduction}

\begin{figure}[t]
  \centering
  \includegraphics[width=\textwidth]{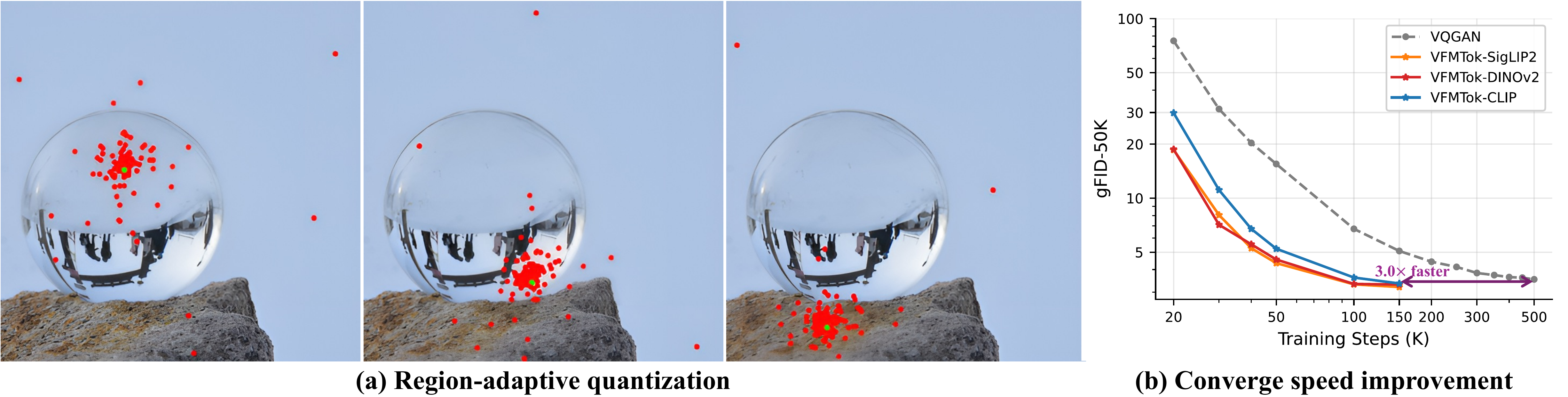}
  \caption{\ours introduces novel features, including: a).\textbf{region-adaptive quantization}— where it adaptively samples regions of similar patterns and extracts their VFM features for quantization; b).\textbf{convergence speed improvement} compared with vanilla VQGAN~\cite{llamagen} for AR image synthesis.}
  \label{fig:pilot2}
  \vspace{-.8cm}
\end{figure}

GPT's success in language generation has spurred interest in autoregressive (AR) image generation~\cite{var, llamagen, show-o}, which relies on visual tokenizers like VQGAN~\cite{vqvae, vqvae2, vqgan,vit-vqgan, llamagen} to map images into compact, discrete latent spaces. However, these tokenizers, typically trained from scratch and optimized for reconstruction, often yield latent spaces rich in low-level details but poor in high-level semantics and laden with redundancy. Such flawed latent spaces not only prolong AR model training (\cref{fig:pilot2}) but also necessitate techniques like classifier-free guidance (CFG) for high-fidelity, class-conditional image generation, which in turn increases inference time.

In parallel, within the field of computer vision, pre-trained vision foundation models such as DINOv2 and CLIP~\cite{dinov2, dinov2reg, clip, siglip, siglip2} have demonstrated strong capabilities in extracting semantically rich and generalizable visual features. Early explorations in diffusion-based image generation---\eg, REPA~\cite{repa}---suggest that the semantic representations learned by these models can facilitate the training of generative models. This leads to a natural and compelling question: \textit{Can the latent features from vision foundation models, originally designed for visual understanding, also serve as robust and structured representations for image reconstruction and generation?}

Recent studies~\cite{vqgan-lc, zhu2024stabilize} have started exploring this direction by leveraging features from vision foundation models to initialize quantizer codebooks~\cite{vqgan-lc, zhu2024stabilize}, augment VQGAN architectures with additional branches~\cite{tokenflow}, or distill these features to guide latent space learning~\cite{dualtoken}. Although these approaches show promise, they typically treat foundation model features as auxiliary components rather than fully capitalizing on their potential as generative priors. As a result, these methods often suffer from inefficiencies and fail to fully utilize the rich semantic information embedded in foundation model features, leaving their generative capabilities largely underexplored.

\textbf{Can VFMs be effective tokenizers?} 
To address this, we initialized the encoder of a VQGAN with different frozen pre-trained foundation models to reconstruct images. Once trained, the tokenizer is integrated on top of an AR model for image synthesis (implementation details depicted in \cref{method:pilot}) As shown in \cref{tbl:prelimiary_exp} (middle rows), our results demonstrate that \textbf{the semantically rich features from these foundation models not only support effective image reconstruction but also achieve generative performance comparable to---or even surpassing---that of a fully trained VQGAN encoder optimized for both reconstruction and generation}. These findings highlight the strong potential of pre-trained vision foundation models to serve as efficient and effective tokenizers for image generation tasks, eliminating the need for extensive encoder training while improving qualities.  


\textbf{Can we improve token efficiency for VFMs?} Building on this pilot study, we are further motivated by the observation that natural images often consist of irregular regions that exhibit recurring visual patterns. For example, as illustrated in \cref{fig:pilot2}(a), the upper portion of the crystal ball exhibits consistent patterns such as texture and transparency; similarly, the moss in the stone possesses similar textural structure. When such images are represented using a regular 2D feature grid extracted from foundation models, this structure-agnostic representation may introduce significant redundancy. Exploiting redundancy within semantically coherent regions offers a promising direction for improving tokenization efficiency. Motivated by this insight, we propose a region-adaptive strategy to refine the latent space that aims to enhance both image reconstruction and generation quality while significantly improving token representation efficiency.

\begin{wraptable}{r}{.63\textwidth}
    \vspace{-1.3em}
    \caption{Pilot study of image reconstruction and generation on ImageNet~\cite{imagenet}. Relative wall-clock inference time for the tokenizer (compared to VFMTok) is reported. L.P. denotes linear probing results on the ImageNet validation set, used to estimate the semantic quality of latent tokens.}
	\label{tbl:prelimiary_exp}
    \vspace{-.5em}
	\centering
    \tablestyle{2.2pt}{1.05}
	\begin{tabular}{l|ccc|ccc|c}
		\toprule
    \multirow{2}{*}{Setup} & \multicolumn{3}{c|}{Image Recon.} & \multicolumn{3}{c|}{AR Generation} & L.P. \\
    & \#Tok. & rFID$\downarrow$ & rIS$\uparrow$ & gFID$\downarrow$ & gIS$\uparrow$ & Time & (\%) \\
    \midrule
    VQGAN~\cite{llamagen} &{576}  & 0.95 & 197.3 & 3.71 & 228.3 & 4.3 & 23.1 \\  \midrule 
    VQGAN (CLIP) & &  1.47 & 182.0  & 3.45 & 221.2 &4.0 & 59.5 \\ 
    VQGAN (SigLIP2) & &  0.96 & 198.4  & 3.39 & 267.8 & 4.0 & 55.5 \\ 
    VQGAN (DINOv2) & \multirow{-3}{*}{576}&  0.99 & 206.3  & 3.34 & 268.6 & 4.0& 56.4 \\ 
    \midrule
    VFMTok (CLIP) &  & 0.99 & 200.1 & 3.40 & 274.7 & 1.0 & 63.9 \\ 
    VFMTok (SigLIP2) &  & 0.94 & \textbf{218.7} & \textbf{3.01} & \textbf{280.8} & 1.0 & \textbf{78.5} \\ 
    VFMTok (DINOv2) & \multirow{-3}{*}{256} & \textbf{0.89} & {215.4} & {3.08} & {274.2} & 1.0 & 69.4 \\ 
    \bottomrule
	\end{tabular}
    \vspace{-1.5em}
\end{wraptable}

\textbf{Our solution and results.} Guided by the preceding experimental analysis and insights, we introduce \ours, an image tokenizer that leverages a frozen pre-trained vision foundation model for adaptive region-level tokenization. \ours is designed to achieve high reconstruction and generation quality with improved token efficiency. Specifically, \ours employs a frozen pre-trained VFM as an encoder to extract multi-level semantic features. A set of learnable anchor queries performs region-level sampling on these features via deformable attention~\cite{deform_detr}, producing region-adaptive tokens that are subsequently quantized into discrete tokens representing the image's latent representation. 
These contextual tokens are then processed by a lightweight Vision Transformer~\cite{dosovitskiy2020image}(ViT) in a BERT-style framework~\cite{bert, mae} with two primary reconstruction objectives. First, the original image pixels are reconstructed after de-quantization using a VQGAN~\cite{llamagen} decoder. Then, the model reconstructs the features from the frozen foundation model itself, allowing \ours to retain the semantic richness and discriminative power of the original representations.
Once trained, \ours enables standard autoregressive Transformers (\eg, Llama~\cite{touvron2023llama}) to generate contextual token sequences, which are decoded back into images via the VQGAN decoder, facilitating high-quality image synthesis with compact and semantically meaningful representations. As shown in \cref{tbl:prelimiary_exp} (bottom rows), VFMTok achieves superior reconstruction and generation performance while using fewer than half the original number of tokens (256 \vs 576). 


Extensive experiments validate that \ours, by combining the representational power of visual foundation models with a novel region-adaptive tokenization strategy based on irregular sampling and learnable anchor queries, enables both high-quality and efficient image reconstruction and autoregressive (AR) generation.
First, \ours achieves superior reconstruction quality and captures richer semantics using significantly fewer tokens compared to prior methods (\eg, 256 \vs 576 in~\cite{llamagen}), resulting in a structured, semantic-aware, and compact latent space. As shown in \cref{tbl:prelimiary_exp}, VFMTok, with only 256 tokens, outperforms other tokenizers using the same VFM encoder by delivering superior reconstruction quality and stronger semantic representation (as indicated by linear probing). Second, the high-quality latent space produced by VFMTok facilitates effective AR training using a simple LLaMA-based model, leading to faster convergence (see \cref{fig:pilot2}(b)) and improved generation performance. Notably, the 1.4B AR model surpasses the performance of LlamaGen-3B despite having fewer parameters and requiring fewer training iterations. The 1.5B advanced AR model achieves a new state-of-the-art with a gFID of \textbf{1.36} on ImageNet~\cite{imagenet} $256\times256$, outperforming widely-used diffusion models. Third, due to the compact token space and the reduced number of tokens, \ours significantly improves the inference speed of AR models (see \cref{tbl:prelimiary_exp}). Moreover, the rich semantic content embedded in the latent tokens enables effective class-conditional image synthesis with high fidelity---without the need for classifier-free guidance---further reducing inference time. 

Our contributions can be summarized as follows:
\begin{itemize}[leftmargin=*]
    \item We demonstrate that frozen vision foundation models---ranging from self-supervised to language-supervised---are effective for image reconstruction and generation. Leveraging their semantic richness enhances the tokenizer and enables AR generation models to converge faster and perform high-fidelity, CFG-free image synthesis, without bells and whistles.
    \item We propose a region-adaptive tokenization framework that effectively leverages inherent redundancies in image regions to achieve compact tokenization. This approach reduces the number of visual tokens while enhancing performance, enabling efficient AR generation without sacrificing quality. 
    \item Extensive experiments validate the effectiveness of our approach in both image reconstruction and AR generation, establishing pre-trained vision foundation models as powerful tokenizers for high-quality and efficient image generation.
    \end{itemize}

%% file: sections/2_related_work.tex
\vspace{-0.4cm}
\section{Related Work}
\label{related_work}

\vspace{-0.1cm}
\noindent\textbf{Image Tokenization using Autoencoders.} Pixel-space images are highly redundant. Autoencoder-based tokenizers~\cite{magvit, magvit2, var, llamagen} create compact latent tokens to reduce redundancy. VQVAEs~\cite{vqvae, vqvae2, kingma2013auto} and their derivatives evolved using adversarial losses~\cite{vqgan}, Transformers~\cite{vit-vqgan}, multistage quantization~\cite{residual-vq, movq}, lookup-free methods~\cite{magvit2, fsq}, and codebook initialization from pre-trained features~\cite{vqgan-lc, zhu2024stabilize}). These 2D tokenizers map features to a static 2D grid, which limits redundancy exploration. Recent 1D tokenizers~\cite{flextok, titok, oneD, semanticist} offer superior compression, reconstruction, and redundancy removal, but often require complex and lengthy training. For example, TiTok~\cite{titok} requires a two-stage process (warming up and fine-tuning) for 200 epochs. Our VFMTok adopts a novel region-adaptive tokenization framework to reduce redundancy. With a simpler training strategy for only 50 epochs, VFMTok exhibits discriminative semantics and excellent generation results.

\vspace{-0.1cm}
\noindent\textbf{Vision Foundation Models.} Vision Foundation Models (VFMs)~\cite{resnet,dino,dinov2, byol, he2019moco, chen2020mocov2,clip,align,siglip,siglip2} aim to learn general, transferable visual representations from large-scale, diverse data. The training of these versatile models has shifted from early supervised approaches to more scalable self-supervised learning~\cite{dino, byol, he2019moco, chen2020mocov2, bert, mae, dinov2, dinov2reg}, which leverages inherent data structures. More recently, language-supervised pre-training~\cite{siglip, align,siglip2} on vast image-text pairs has enabled VFMs to learn rich, semantically grounded representations. Pre-trained VFMs serve as powerful backbones for a wide array of downstream tasks. In this work, we utilize pre-trained VFMs directly as image tokenizers for AR image generation, surpassing other methods~\cite{vqgan-lc, zhu2024stabilize} with superior performance. Furthermore, using VFMs as tokenizers enables the removal of classifier-free guidance.

\vspace{-0.1cm}
\noindent\textbf{Autoregressive Image Generation.} GPT-style Transformers~\cite{gpt, chen2020generative, pixelgpt, llamagen, residual-vq, var} have spurred interest in autoregressive (AR) image generation, which predicts visual token sequences. While early AR models operated in pixel space~\cite{chen2020generative, pixelgpt}, current methods~\cite{residual-vq, vit-vqgan, var, llamagen} generate discrete latent tokens via next-token prediction, then decode them to pixels using a tokenizer's decoder~\cite{vqvae, vqvae2, vit-vqgan, vqgan}. To improve the generation quality, recent works~\cite{var, mar, show-o} add bidirectional attention (\eg, VAR's next-scale prediction~\cite{var}, MAR's BERT-style framework~\cite{mar}, Show-o's hybrid attention~\cite{show-o}). These innovations, however, complicate designing universal, multi-modal Transformers adhering to next-token prediction. Instead, our {\ours} enables standard AR transformers to generate contextual token sequences for subsequent decoding, eliminating complex structural modifications. 


%% file: sections/3_approach.tex
\section{Method}
In this section, we first provide preliminary background on quantized image tokenizers. We then present our pilot studies exploring the use of vision foundation models for tokenization. Finally, we introduce \ours, a novel tokenizer built upon frozen vision foundation models, incorporating region-adaptive strategies to enhance both the efficiency and effectiveness of the tokenization process. 

\subsection{Preliminary}
\label{method:preliminary}

\noindent\textbf{Quantized Image Tokenizer.} \label{vq-procedure}
To apply autoregressive modeling to visual generation, existing methods~\cite{vit-vqgan,yu2022scaling,llamagen,var} necessitate an image tokenizer to map a 2D image into discrete token sequences for AR generation. To achieve this, quantized autoencoders, such as VQVAEs \cite{vqvae, vqvae2,vqgan, vit-vqgan,vqgan-lc,var,llamagen}, are widely used. Typically, an image tokenizer consists of an encoder $\mathcal{E}(\cdot)$, a quantizer $\mathcal{VQ}(\cdot)$, and a decoder $\mathcal{D}(\cdot)$.  Given an input image $\text{I} \in \mathbb{R}^{H \times W \times 3}$, the encoder $\mathcal{E}(\cdot)$ first convert an image into patch embeddings $Z_{2D} \in \mathbb{R}^{\frac{H}{f} \times \frac{W}{f} \times \text{D}}$ with spatial down-sampling factor $f$. Then, $Z_{2D}$ is fed into the quantizer $\mathcal{VQ}(\cdot)$ that includes a learnable codebook $\mathbb{C} \in \mathbb{R}^{N \times D}$ with $N$ vectors. Each feature vector ${z_i} \in \mathbb{R}^{D}$ is mapped into its nearest vector ${c_i} \in \mathbb{R}^{D}$ in the codebook $\mathbb{C}$. 
\begin{equation}
\begin{split}
{Z_{2D}} &= \mathcal{E}(\text{I}) \,, \\
\mathcal{VQ}({z_i}) = {c_i}, \quad \text{where} \quad {i} &= \mathop{\arg\min}\limits_{{j} \in \{1, 2, ..., N\}} \Vert {z_i} - {c_j} \Vert_2 \,,
\end{split}\label{equ:vq}
\end{equation}
where $H$ and $W$ denote the input image’s height and width, respectively. $D$ depicts the latent feature dimension.  Once discrete tokens are acquired, they can be de-quantized into corresponding code and converted back to image pixels by the decoder $\mathcal{D}(\cdot)$, as depicted in \cref{equ:dec}.
\begin{equation}
\label{equ:dec}
\mathbf{\hat{I}} = \mathcal{D}(\mathcal{VQ}({Z_{2D}})) \,.
\end{equation}
To optimize the codebook, the training objective is $\mathbf{\mathcal{L}_{vq}}=\sum{\|\mathbf{sg}(z_i) -{c_i}\|^{2}_{2} + {\beta} \cdot \| \mathbf{sg}({c_i}) - {z_i}\|^{2}_{2}} $, where $\mathbf{sg}(\cdot)$ is a stop-gradient function~\cite{straightsthrough,vqvae}. The second term is a commitment loss weighted by $\beta$ to align extracted features with codebook vectors. For image reconstruction, the loss function is $\mathcal{L}_{AE} = \mathcal{L}_2(\text{I}, \hat{\text{I}}) + \mathcal{L}_{P}(\text{I}, \hat{\text{I}}) + \lambda_{G} \cdot \mathcal{L}_{G}(\hat{\text{I}})$, where $\mathcal{L}_2$ is a pixel-wise reconstruction loss, $\mathcal{L}_{P}$ is perceptual loss from LPIPS \cite{percetualloss}, and $\mathcal{L}_{G}$ is adversarial loss from PatchGAN \cite{patchgan} with weight $\lambda_{G}$.

\subsection{Pilot Study: Pre-trained Vision Foundation Models as Tokenizers for AR Generation}
\label{method:pilot}

To assess whether a pre-trained VFM can serve as a tokenizer for image reconstruction and benefit image generation, we performed a pilot study. In our setup, we extract the final 2D grid features from images of size $336\times336$ using a frozen VFM, such as DINOv2, CLIP, and SigLIP2. These features, after quantization, are fed into a VQGAN~\cite{llamagen} decoder for image reconstruction. Once trained, the tokenizer is integrated on top of a Llama-based AR model for image synthesis. Additionally, the training duration for VQGANs and AR models is 50 and 100 epochs, respectively.

As \cref{tbl:prelimiary_exp} illustrates, directly using features from pre-trained VFMs yields decent image reconstruction and generation performance compared to vanilla VQGANs. Notably, these VFM-based tokenizers consistently exhibit stronger semantic representation capabilities (as indicated by the linear probing experiment in \cref{tbl:prelimiary_exp}). For instance, VQGAN (SigLIP2) achieves reconstruction performance on par with vanilla VQGAN, while exhibiting better semantic representation and superior generation quality. Nevertheless, variations in image reconstruction and generation quality arise when different VFMs are used to initialize the tokenizer's encoder. Specifically, VQGAN (DINOv2) and VQGAN (SigLIP2) demonstrate similar reconstruction and generation quality, both outperforming vanilla VQGAN, while the reconstruction quality of VQGAN (CLIP) trails that of vanilla VQGAN. One contributing factor is that different learning objectives used to train VFMs influence their ability to extract detailed and semantic features from images, thereby affecting downstream image reconstruction and generation quality. As evidence, both DINOv2~\cite{dinov2reg} and SigLIP2~\cite{siglip2} employed a masked prediction objective to optimize their VFMs, whereas CLIP~\cite{clip} did not.

\subsection{VFMTok}

Building upon the semantically rich features provided by vision foundation models---typically structured as regular 2D grids---we introduce VFMTok, a region-adaptive tokenizer that identifies semantically coherent, irregular local regions to produce region-adaptive tokens. These tokens are sequentially quantized for decoding, with tailored learning objectives to enhance performance. In the following, we detail the architecture of VFMTok, including its region-adaptive token generation module and dedicated decoder for both image and feature reconstruction. We further describe the training objectives, which combine a pixel-level reconstruction loss for image synthesis with a feature reconstruction loss that preserves the semantic content of the foundation model’s representations. 



\begin{figure}[thbp]
  \centering
  \includegraphics[scale=0.43]{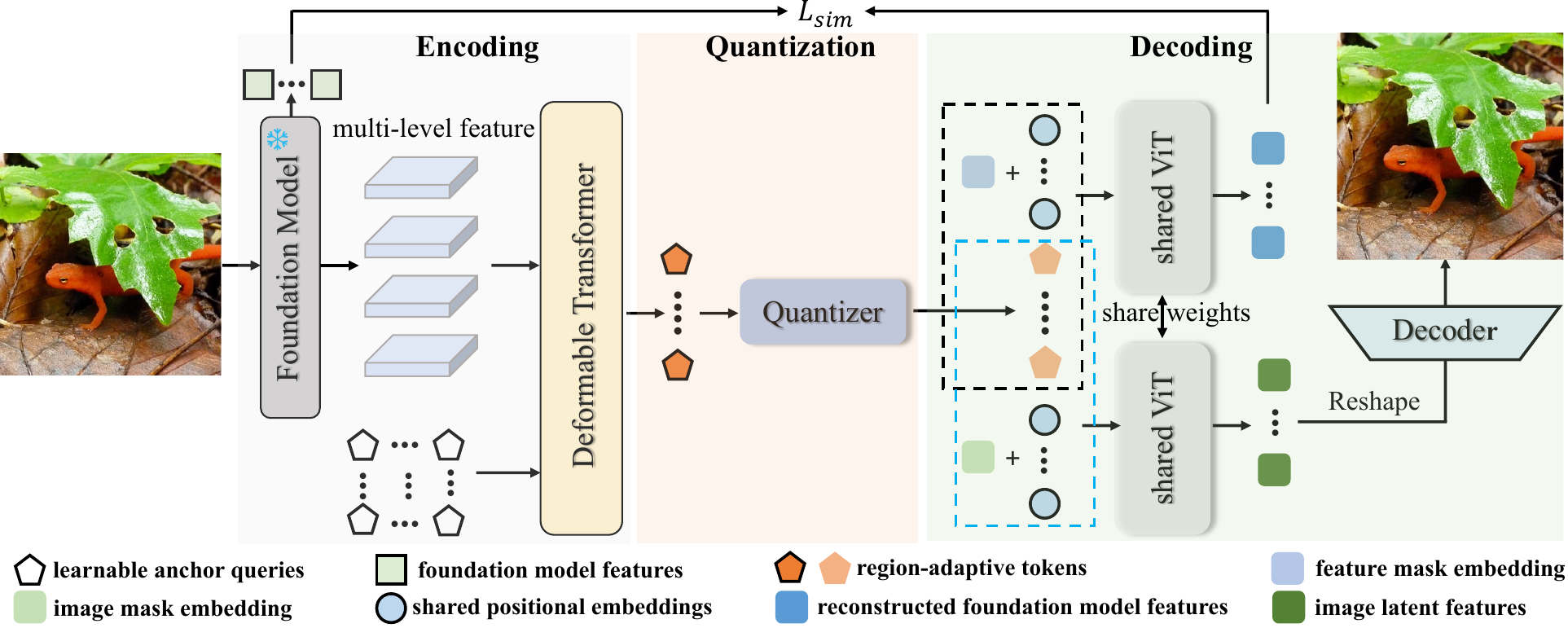}
  \caption{The framework of \ours. \ours utilizes a frozen VFM to extract multi-level image features. A deformable Transformer then processes these features with learnable grid queries to generate region-adaptive tokens. After quantization, these tokens are fed into a shared ViT for dual reconstruction: 1) VFM features, targeting similarity with the VFM's last-layer outputs, and 2) image latent features, which are reshaped to a 2D grid and decoded into pixels.}
  \label{fig:framework}
\end{figure}

\noindent\textbf{Region-adaptive Token Generation.}
Following our pilot study, we utilize a frozen pre-trained vision foundation model (VFM) to encode an input image $I$ into latent embeddings  $\mathcal{F}$. Since features extracted from VFMs contain rich details in shallower layers and high-level semantics in deeper layers~\cite{opmp, fpn, retinanet}---both of which are critical for image reconstruction---we extract multi-level features $\mathcal{F_\text{m}}$ from the VFM. These multi-level features are then projected to a uniform embedding dimension using a two-layer MLP.
Next, {as shown in Fig.~\ref{fig:framework}}, based on the multi-level features $\mathcal{F_\text{m}}$, we introduce a region-adaptive sampling mechanism using deformable cross-attention layers~\cite{deformable_conv, deform_detr}. A set of learnable anchor queries, initialized as a 2D grid, are iteratively refined through multiple deformable attention layers. In each layer, an anchor query predicts sampling offsets for each VFM feature level via a linear layer, enabling sampling from irregular, data-dependent positions. These sampled features are then weighted using attention scores---computed through another linear layer---and aggregated to update the query.
Through this process, the anchor queries are progressively refined to capture semantically coherent, region-specific information. The final refined queries are referred to as region-adaptive tokens $Z_r$, which are subsequently quantized into discrete tokens $\Tilde{Z}_r$. Compared to a fixed 2D feature grid, VFMTok adaptively aggregates features from semantically coherent, irregular regions. This substantially reduces redundancy, enabling the use of fewer tokens while maintaining superior image reconstruction and generation performance. As shown in Tab.~\ref{tbl:prelimiary_exp}, just 256 semantically rich tokens from VFMTok are sufficient to achieve high-fidelity reconstruction and generation.

\noindent\textbf{Vector Quantization.}
\label{sec:quantization}
Once the continuous region-adaptive tokens $Z_r$ are obtained, a quantizer $Q_c(\cdot)$ is applied to discretize them into region-adaptive discrete tokens $\Tilde{Z}_r$. Given that the design of the codebook plays a critical role in the performance of an image tokenizer, we follow the practices in~\cite{vit-vqgan, llamagen} by applying $\ell_2$-normalization to the codebook vectors. Additionally, we adopt a low-dimensional embedding space with a large codebook size to enhance both reconstruction quality and codebook utilization following~\cite{vit-vqgan, llamagen}.

\noindent\textbf{Decoder of VFMTok  for Image and VFM Feature Reconstruction.}
After de-quantization, the region-adaptive tokens $\Tilde{Z}_r$ are used for image reconstruction. Since these tokens represent irregular, region-level features, decoding them into a regular 2D image grid requires alignment. To achieve this, we introduce a set of mask tokens $M_\text{I}$, representing a 2D feature map of size $H_m \times W_m$ with channel dimension $C$. The mask tokens are initialized by replicating a single learnable token  $H_m \times W_m$ times. Position embeddings $E$, encoding spatial locations, are then added to form position-aware masked tokens.
Next, the de-quantized region-adaptive tokens $\Tilde{Z}$ are concatenated with  $M_\text{I}$, and the combined sequence is processed by a lightweight Transformer $\mathcal{E}_\text{ViT}(\cdot)$, which propagates information from the region-adaptive tokens to the masked image tokens. This Transformer employs \textit{causal} self-attention, aligning its latent space with the structure of autoregressive models. Following DINOv2~\cite{dinov2reg}, we further enrich the input sequence by appending a $\verb+[CLS]+$ token and several register tokens to improve representation learning and capture global context---though these are not used for reconstruction. The output of this Transformer is a refined set of mask tokens $\mathcal{F}_I$ representing a regular 2D grid structure. These are reshaped into a spatial grid and passed into a decoder $\mathcal{D}(\cdot)$ to reconstruct the image.

To preserve the semantic integrity of the VFMTok tokens, we also reconstruct high-level features (specifically, from the final layer) of the vision foundation model (VFM). This process mirrors image reconstruction: a new set of mask tokens $\textbf{M}_f$ is initialized and augmented with positional embeddings $E$, shared with those used in image reconstruction.  The concatenation of $Z_r$ and $\textbf{M}_f$ is then processed by the same shared Transformer $\mathcal{E}_\text{ViT}(\cdot)$ to produce  $\mathcal{F}_\text{P}$, the reconstructed high-level VFM feature map. 
By sharing $\mathcal{E}_\text{ViT}(\cdot)$ between image and feature reconstruction, we reduce the model's parameter footprint while ensuring the semantic fidelity of the latent tokens.  
Note that the VFM feature reconstruction is only applied during tokenizer training.

\noindent\textbf{Training Objective.}
For tokenizer optimization, we follow the training objectives of VQGAN~\cite{vqgan, llamagen}, with one key modification: we replace its original discriminator with a pre-trained DINOv1-S~\cite{dino} model. This substitution provides adversarial training guidance in a more semantically meaningful way compared to conventional discriminators such as PatchGAN~\cite{patchgan}, and we find it consistently improves reconstruction quality.
In addition to image reconstruction, we incorporate a feature reconstruction objective by computing the cosine similarity loss between the reconstructed features and the corresponding frozen features from the pre-trained vision foundation model (VFM). The overall training loss is defined as: $\mathcal{L} = \alpha \cdot \mathcal{L}_\text{AE} + \lambda \cdot \mathcal{L}_{\text{sim}}$, where $\mathcal{L}_\text{AE}$ denotes the image reconstruction loss and $\mathcal{L}_{\text{sim}}$ is the feature reconstruction loss. In our experiments, we set both $\alpha$ and $\lambda$ to 1.

\subsection {Autoregressive Image Generation}
Once VFMTok is trained, the optimized discrete region-adaptive tokens $\Tilde{Z}_r$ can be integrated into an autoregressive (AR) Transformer, where they are generated sequentially via a next-token prediction mechanism, conditioned on a class or text embedding $c$. The generated tokens are then passed through the Transformer encoder $\mathcal{E}_{\text{ViT}}(\cdot)$ to produce latent image features $\mathcal{F}_{\text{I}}$, which are subsequently decoded into images using the decoder $\mathcal{D}(\cdot)$.
In the AR model, the region-adaptive tokens $\Tilde{Z}_r$ are augmented with positional embeddings—specifically 2D Rotary Position Embeddings (RoPE)~\cite{rope}---to better capture their spatial locality and structure.

%% file: sections/4_experiments.tex
\section{Experiment}
\subsection{Setup} \label{sec:setup}

\noindent\textbf{Image Tokenizer.}
In the main experiment, we initialize the encoder of VFMTok with a frozen pre-trained DINOv2-L~\cite{dinov2reg}. Considering its composition of 24 Transformer layers, we extract features from the 6th, 12th, 18th, and 24th layers to create multi-level features. Consistent with \cite{llamagen, titok}, we set the codebook vector dimension of the quantizer to 12 with a codebook size of 16384, to achieve a better reconstruction quality and efficient codebook utilization. Meanwhile, {\ours} utilizes 256 tokens to represent an image. Besides, the depth of the Transformer is set to 6 (following \cite{deform_detr}). The model is trained on the ImageNet~\cite{imagenet} training set and evaluated on its validation set.

Given that the resolution of vision foundation models (VFMs)~\cite{dinov2reg, clip,align, siglip,siglip2} is typically $336\times336$, while \ours represents images with fixed 256 tokens by default, it's comparable to vanilla tokenizers~\cite{vqgan,vqgan-lc, llamagen}. Thus, we train the tokenizer on $336\times336$ images. Except this, we keep the training settings unchanged as LlamaGen~\cite{llamagen}. During evaluation, the reconstructed images of $336\times336$ are resized to $256\times256$ for evaluation, which is consistent with the evaluation procedure in LlamaGen~\cite{llamagen}.



\noindent\textbf{Class-conditional Autoregressive Image Generation.} 
Following the generation procedure in LlamaGen~\cite{llamagen}, the AR models first generate images of $336\times336$ and then resize them to $256\times256$ for evaluation. Considering computational costs, we set the training duration based on the number of models' parameters. Models with fewer than 1B parameters are trained for 300 epochs, while the remaining models are trained for 200 epochs. Beyond the resolution and training duration, all models are trained with the same settings as LlamaGen~\cite{llamagen}. Furthermore, we also incorporated the same VFMTok into the RAR~\cite{rar} autoregressive generation framework, with all training settings remaining consistent with RAR~\cite{rar}. Additionally, in our experiments, AR generation is conducted with both classifier-free guidance (CFG) and a CFG-free protocol. 

\noindent\textbf{Evaluation metrics.} 
To evaluate image generation performance, we use Fréchet inception distance (FID)~\cite{fid} and Inception Score (IS)~\cite{is} as the main metrics to measure the generation quality of different models. In addition, sFID, Precision, and Recall~\cite{prec_recall} are also reported following~\cite{llamagen}.
\begin{wraptable}{r}{0.65\textwidth}
    \caption{Comparison with other image tokenizers. $^\text{oim.}$ indicates trained on OpenImages~\cite{openimage}. $\mathcal{Q}_{c}$/$\mathcal{Q}_{P}$ denotes the codebook usage in contextual and patch-level quantizers, respectively. 
    }
	\label{tbl:tok_comparison2}
	\centering
    \tablestyle{2.pt}{1.05}
	\begin{tabular}{l|c|ccc|cc|cc}
	\toprule
    \multirow{2}{*}{Method} &  & \multicolumn{3}{c|}{Tokenizer Setup} & \multicolumn{2}{c|}{Image Recon.} & \multicolumn{2}{c}{Usage (\%)$\uparrow$} \\
     & $\textit{f}$ & Size & Dim. & \#Tok. & rFID$\downarrow$ & rIS$\uparrow$  & $\mathcal{Q}_{C}$ & $\mathcal{Q}_{P}$ \\
    \midrule
    TiTok~\cite{titok} & -- & 8192 & 64 & 256 & 1.05 & 191.5 
    & 100 & -- \\
    ImageFolder~\cite{imagefolder} & -- & 32768 & 32 & 286 & \textbf{0.69} & 201.5 
    & 100 & -- \\
    \midrule
    $\text{VQGAN}^{\text{oim.}}$~\cite{vqgan} & \multirow{4}{*}{8} & 256 & 4 & \multirow{4}{*}{1024} & 1.44 & -- & -- & -- \\
    VQGAN~\cite{vqgan} & ~ & 8192 & 256 & ~ & 1.49 & -- & -- & -- \\
    ViT-VQGAN~\cite{vit-vqgan} & ~ & 8192 & 32 & ~ & 1.28 & 192.3 & -- & 95.0  \\
    $\text{VQGAN}^{\text{oim.}}$~\cite{vqgan} & ~ & 16384 & 4 & ~ & 1.19 & -- &  -- & --\\
    \midrule
    VQGAN~\cite{vqgan} & \multirow{3}{*}{16} & \multirow{2}{*}{1024} & \multirow{2}{*}{256} & \multirow{2}{*}{256} & 7.94 & -- & --  & -- \\
    MaskGiT~\cite{maskgit} & ~ & ~ & ~ & ~ & 2.28 & -- & -- & -- \\
    VAR~\cite{var} &  & 4096 & 32 & 680 & {0.92} & 196.0  & -- & 100 \\
    \midrule
    RQ-VAE~\cite{residual-vq} & 32 & 16384 & 256 & 1024 & 1.83  & -- & -- & -- \\
    \midrule
    VQGAN~\cite{vqgan} & \multirow{3}{*}{16} & \multirow{3}{*}{16384} & 256 & 256 & 4.98 & --  & --  & --\\
    VQGAN~\cite{llamagen} & ~ & ~ & \multirow{2}{*}{8} & 441 & 1.21 & 189.1  & -- & 99.2 \\
    VQGAN~\cite{llamagen} & ~ & ~ &  & 576 & {0.95} & 197.3  & -- & 99.7 \\
    \midrule
    \textbf{VFMTok~(\textit{Ours})}  & \multirow{1}{*}{--} & \multirow{1}{*}{16384} & \multirow{1}{*}{12} & \multirow{1}{*}{256} & {0.89} & \textbf{215.4}  & 100 & -- \\
    \bottomrule
	\end{tabular}
    \vspace{-1.cm}
\end{wraptable}

\vspace{-1.1cm}
\subsection{Main Results}

\vspace{-0.3cm}
\noindent\textbf{Image Reconstruction.} We compare {\ours} against representative 2D image tokenizers, VQGAN~\cite{vqgan}, MaskGiT~\cite{maskgit}, ViT-VQGAN~\cite{vit-vqgan}, and 1D tokenizer, TiTok~\cite{titok}. As shown in Tab.~\ref{tbl:tok_comparison2}, our tokenizer represents an image with just 256 tokens, considerably fewer than some counterparts. For instance, the VQGAN variant LlamaGen~\cite{llamagen} uses 576 tokens, while VQGAN~\cite{vqgan} and ViT-VQGAN~\cite{vit-vqgan} even utilize up to 1024 tokens. Despite this efficiency, {\ours} achieves a strong rFID of \textbf{0.89}, and further demonstrates 100\% utilization of the codebook.

The rIS score of \textbf{215.4} achieved by \ours significantly outperforms other methods, \eg, TiTok~\cite{titok} and the VQGAN series. The rIS metric quantifies the KL-divergence between the original label distribution and the logit distribution of reconstructed images after softmax normalization, thereby measuring the semantic consistency between reconstructed and original images. The higher rIS confirms {\ours} is more effective at preserving semantic consistency during reconstruction.

\begin{table*}[ht]
	\caption{Class-conditional image generation quality estimated on ImageNet~\cite{imagenet} validation benchmark. $^{\dagger}$ indicates it is implemented by us, and `-re' indicates using rejection sampling.}
	\label{tbl:distict-gen}
	\centering
    \tablestyle{1.pt}{1.05}
	\begin{tabular}{l|l|cccccccc|ccccc}
    \toprule
\multirow{2}{*}{Type} & \multirow{2}{*}{{Method}} & \multirow{2}{*}{\#Epoch} & \multirow{2}{*}{{\#Para.}} & \multirow{2}{*}{{\#Tok.}} & \multicolumn{5}{c|}{{Generation w/ CFG}} & \multicolumn{5}{c}{{Generation w/o CFG}} \\
 & &  & & & {gFID} & {sFID} & {gIS} & {Pre.} & {Rec.} & {gFID} & {sFID} & {gIS} & {Pre.} & {Rec.}  \\
  \midrule
\multirow{4}{*}{Diff.} & MaskDiT~\cite{maskDiT}  & 1600 & 675M & \multirow{4}{*}{--} & 2.28 & 5.67 & 276.6 & 0.80 & 0.61 & 5.69 & 10.34 & 177.9 & 0.74 & 0.60 \\
&  DiT~\cite{peebles2023scalable}  & 1600 & 675M &   & 2.27 & 4.60 & 278.2 & 0.83 & 0.57 & 9.62 & 6.85 & 121.5 & 0.67 & 0.67 \\
&  SiT~\cite{sit} & 1600 & 675M &  &  2.06 & {4.50} & 270.3 & 0.82 & 0.59 & 8.61 & 6.32 & 131.7 & 0.68 & 0.67 \\
&  FasterDIT~\cite{fasterdit} & 400 & 675M & & {2.03} & 4.63 & 264.0 & 0.81 & 0.60 & 7.91 & 5.45 & 131.3 & 0.67 & {0.69}  \\
  \midrule
\multirow{2}{*}{Mask.} & MaskGiT~\cite{maskgit} & \multirow{2}{*}{555} & \multirow{2}{*}{227M} & \multirow{2}{*}{256} & -- & -- & -- & -- & -- & 6.18 & -- & 182.1 &  0.80 & 0.51  \\
 & MaskGiT-re  &  &  &  & 4.02 & -- & {355.6} &  -- & -- & -- & -- & -- & -- & -- \\
  \midrule
\multirow{20}{*}{AR}&  VAR~\cite{var} & 350 & 310M & 680 & 3.30 & -- & 274.4 & 0.84 & 0.51 & -- & -- & -- & -- & -- \\
  \cmidrule{2-15}
 & TiTok-B$^{\dagger}$~\cite{titok} & \multirow{2}{*}{300} & 111M & \multirow{2}{*}{256} & 6.76 & 7.82 & 175.3 & {0.85} & 0.43 & 19.6 & 57.9 & 7.54 & 0.64 & 0.60 \\
 & TiTok-L$^{\dagger}$~\cite{titok}  &  & 343M &  & 4.03 & 6.93 & 219.5 & 0.84 & 0.52 & 11.4 & 88.8 & 7.14 & 0.68 & 0.64  \\
  \cmidrule{2-15}
 & LlamaGen-B & \multirow{4}{*}{300} & 111M & \multirow{4}{*}{576} & 6.09 & 7.24 & 182.5 & {0.85} & 0.42 & 32.2 & 11.84 & 39.9 & 0.57 & 0.61\\
&  LlamaGen-L  &  & 343M &  & 3.07 & 6.09 & 256.1 & 0.83 & 0.52 & 19.1 & 8.67 & 64.3 & 0.61 &  0.67 \\  
 & LlamaGen-XXL  &  & 1.4B &  & 2.34 & 6.00 & 253.9 & 0.81 & 0.60 & 14.6 & 8.69 & 86.3 & 0.63 & 0.68 \\
 & LlamaGen-3B  &  & 3.1B &  & 2.19 & 5.97 & 263.3 & 0.82 & 0.58 & 9.38 & 8.24 & 112.9 & 0.69 & 0.67 \\
 \cmidrule{2-15}
  & RAR-L~\cite{rar} & \multirow{3}{*}{400}  & 461M & \multirow{3}{*}{256} & 1.70 & -- & 299.5 & 0.82 & 0.58 & 6.72 & 5.56 & 129.2 & 0.74 & 0.61\\
  & RAR-XL~\cite{rar} &  & 955M &  & 1.50 & -- & 306.9 & 0.80 & 0.62 & 4.62 & 5.27 & 158.3 & 0.77 & 0.62\\
  & RAR-XXL~\cite{rar} &  & 1.5B &  & 1.48 & -- & 326.0 & 0.80 & 0.63 & 3.85 & 5.18 & 176.2 & 0.78 & 0.61\\
  \cmidrule{2-15}
 & \cellcolor{cyan!10}VFMTok-B & \cellcolor{cyan!10} & \cellcolor{cyan!10}111M & \cellcolor{cyan!10} & \cellcolor{cyan!10}3.43 & \cellcolor{cyan!10}5.88 & \cellcolor{cyan!10}252.2 & \cellcolor{cyan!10}\textbf{0.85} & \cellcolor{cyan!10}0.53 & \cellcolor{cyan!10}3.09 & \cellcolor{cyan!10}5.67 & \cellcolor{cyan!10}173.6 &  \cellcolor{cyan!10}0.80 & \cellcolor{cyan!10}0.58 \\
 & \cellcolor{cyan!10}VFMTok-L & \cellcolor{cyan!10}\multirow{-2}{*}{300} & \cellcolor{cyan!10}343M & \cellcolor{cyan!10} & \cellcolor{cyan!10}2.75 & \cellcolor{cyan!10}5.58 & \cellcolor{cyan!10}278.8 & \cellcolor{cyan!10}0.84 & \cellcolor{cyan!10}0.57 & \cellcolor{cyan!10}2.11 & \cellcolor{cyan!10}5.46 & \cellcolor{cyan!10}230.1 & \cellcolor{cyan!10}{0.82} & \cellcolor{cyan!10}0.60 \\ 
 & \cellcolor{cyan!10}VFMTok-XXL & \cellcolor{cyan!10}200 & \cellcolor{cyan!10}1.4B & \cellcolor{cyan!10} & \cellcolor{cyan!10}2.19 & \cellcolor{cyan!10}5.53 & \cellcolor{cyan!10}278.0 & \cellcolor{cyan!10}0.83 & \cellcolor{cyan!10}0.60 & \cellcolor{cyan!10}{1.95} & \cellcolor{cyan!10}5.65 & \cellcolor{cyan!10}259.3 & \cellcolor{cyan!10}{0.82} & \cellcolor{cyan!10}{0.62} \\
& \cellcolor{cyan!10}VFMTok-3B & \cellcolor{cyan!10}200 & \cellcolor{cyan!10}3.1B & \cellcolor{cyan!10}\multirow{-4}{*}{256} & \cellcolor{cyan!10}{2.07} & \cellcolor{cyan!10}6.23 & \cellcolor{cyan!10}280.4 & \cellcolor{cyan!10}0.81 & \cellcolor{cyan!10}\textbf{0.62} & \cellcolor{cyan!10}2.04 & \cellcolor{cyan!10}5.43 & \cellcolor{cyan!10}267.6 & \cellcolor{cyan!10}{0.82} & \cellcolor{cyan!10}0.61  \\
 \cmidrule{2-15}
   &\cellcolor{cyan!10}RAR-L(VFMTok) & \cellcolor{cyan!10} & \cellcolor{cyan!10}461M & \cellcolor{cyan!10} & \cellcolor{cyan!10}1.44 & \cellcolor{cyan!10}6.03 & \cellcolor{cyan!10}{312.8} & \cellcolor{cyan!10}0.78 & \cellcolor{cyan!10}0.66 & \cellcolor{cyan!10}2.02 & \cellcolor{cyan!10}5.51 & \cellcolor{cyan!10}210.4 & \cellcolor{cyan!10}0.79 & \cellcolor{cyan!10}0.63\\
   & \cellcolor{cyan!10}RAR-XL(VFMTok) & \cellcolor{cyan!10} & \cellcolor{cyan!10}955M & \cellcolor{cyan!10}  & \cellcolor{cyan!10}1.38 & \cellcolor{cyan!10}5.86 & \cellcolor{cyan!10}310.2 & \cellcolor{cyan!10}0.78 & \cellcolor{cyan!10}0.65 & \cellcolor{cyan!10}1.74 & \cellcolor{cyan!10}\textbf{5.33} & \cellcolor{cyan!10}233.0 & \cellcolor{cyan!10}0.80 & \cellcolor{cyan!10}0.63\\
   & \cellcolor{cyan!10}RAR-XL(VFMTok) & \cellcolor{cyan!10}\multirow{-3}{*}{400}  & \cellcolor{cyan!10}1.5B & \cellcolor{cyan!10}\multirow{-3}{*}{256}  & \cellcolor{cyan!10}\textbf{1.36} & \cellcolor{cyan!10}5.85 & \cellcolor{cyan!10}301.3 & \cellcolor{cyan!10}0.78 & \cellcolor{cyan!10}{0.66} & \cellcolor{cyan!10}\textbf{1.65} & \cellcolor{cyan!10}5.55 & \cellcolor{cyan!10}253.7 & \cellcolor{cyan!10}0.80 & \cellcolor{cyan!10}0.63\\
\cmidrule{2-15}
& VFMTok-L(SigLIP2) & 300 & 343M & \multirow{3}{*}{256} & 2.69 & \textbf{5.31} & 273.4 & 0.84 &  0.56 & 2.11 & {5.39} & 225.6 & 0.81 & 0.60 \\ 
& VFMTok-XXL(SigLIP2) & 200 & 1.4B & & 2.16 & 5.45 & 272.0 & 0.83 & 0.60 & 1.98 & 5.53 & 265.3 & {0.82} & 0.62 \\
& VFMTok-2B(SigLIP2) & 200 & 2.2B & & 2.17 & 5.43 & 281.4& 0.83 & 0.60 & 1.98 & 5.41 & \textbf{269.7} & {0.82} & 0.62 \\
  \bottomrule
	\end{tabular}
\vspace{-.7cm}
\end{table*}

\noindent\textbf{Class-conditional Image Generation.} We evaluate VFMTok on vanilla autoregressive models -- LlamaGen~\cite{llamagen}, and advanced generative model -- RAR~\cite{rar} with different scales by conducting  $256\times256$ class-conditional image generation task on ImageNet~\cite{imagenet}, where comparing them with the mainstream generation models, including diffusion models (Diff.)~\cite{maskDiT, peebles2023scalable, sit, fasterdit}, BERT-style masked-prediction models (Mask.)~\cite{maskgit}, and AR generation models (AR)~\cite{titok, vqgan, vit-vqgan, residual-vq, vqvae2, llamagen}.

As shown in \cref{tbl:distict-gen}, our models exhibit competitive performance across all metrics compared to mainstream image generation models. Notably,  \ours beats BERT-style models~\cite{maskgit} in terms of gFID without the requirement of complicated sampling tuning. With comparable or even fewer parameters, our method surpasses most AR generative models~\cite{titok, vqgan, vit-vqgan, residual-vq, vqvae2} in both gFID and gIS metrics. Under the same training setting, \ours surpasses LlamaGen~\cite{llamagen} by significant gFID gains and notable gIS improvements. Specifically, \ours-B outperforms LlamaGen-B~\cite{llamagen} with gains of \textbf{2.56} in gFID and \textbf{69.7} in gIS. Besides, our \ours-L model achieves a gFID of \textbf{2.75} at 300 epochs, also obtaining a gain of \textbf{22.7} in gIS. Notably, when compared with LlamaGen-3B with 3B parameters, our {\ours}-XXL achieves even better generation performance with less than half the number of parameters and fewer training iterations. Futhermore, when VFMTok is incorporated into RAR~\cite{rar}, it achieves a generative performance with gFID of~\textbf{1.36}, which is the state-of-the-art generation performance at present.
Additionally, class-conditional image generation results are visualized in the Appendix.

Furthermore, we conducted experiments by \textbf{removing classifier-free guidance (CFG)}. Remarkably, the generation results without CFG show that most evaluation metrics—such as sFID, Precision, and Recall—remain comparable to those obtained with CFG. While gIS experiences a slight decline, gFID improves compared to its CFG-enabled counterpart. Similar trends are observed when VFMTok's encoder is replaced with other frozen pre-trained vision foundation models like SigLIP2~\cite{siglip2}.
These results demonstrate that our method supports high-fidelity autoregressive image generation even without CFG, which significantly accelerates inference. In contrast, baseline methods suffer substantial performance degradation without CFG---for example, LlamaGen-3B model sees gFID worsen to \textbf{9.38}, whereas our 1.4B model VFMTok-XXL achieves a gFID of \textbf{1.95} without CFG.

\subsection{Ablation Study and More Analysis}


\noindent\textbf{Component study.} 
To assess the contribution of each proposed component to image reconstruction and synthesis, we conduct a step-by-step component analysis using a baseline tokenizer built on vanilla VQGAN~\cite{llamagen}. We incrementally add the following components: (1) replace the VQGAN encoder with a frozen pre-trained foundation model (DINOv2-L~\cite{dinov2reg}); (2) introduce learnable queries and a deformable attention for region-adaptive tokenization, using only single-level features from the final layer; (3) incorporate multi-level features to enrich representations with both low-level detail and high-level semantics; and (4) add a feature reconstruction objective based on pre-trained VFM outputs.
After training each tokenizer, we integrate it with our AR generation model, \ours-L, for autoregressive image synthesis. Both the tokenizer and AR model are trained for 50 epochs. Additionally, we perform linear probing on the $\verb+[CLS]+$ token, following the MAE~\cite{mae} protocol.


 As shown in \cref{tbl:ablation_queries}, replacing VQGAN’s encoder with a frozen pre-trained vision foundation model yields reconstruction and generation performance on par with a VQGAN trained specifically for visual reconstruction using 576 tokens. This substitution also significantly enhances the semantic quality of the tokenizer’s representations. To further improve token efficiency, we introduce region-adaptive tokenization using deformable attention to exploit the spatial redundancy inherent in regular 2D grid features. This reduces the number of visual tokens to 256. However, this performance gain comes at a cost: reconstruction and generation quality degrade slightly due to two factors: (1) fewer visual tokens limit representational capacity, and (2) the absence of explicit supervision hinders the effective optimization of the region-adaptive tokens. To address this, we incorporate multi-level feature extraction, which improves the reconstruction capability by leveraging both low- and high-level information. However, without additional guidance, the semantic consistency of the learned tokens may still degrade. Finally, we introduce a pre-trained feature reconstruction objective, which significantly boosts both image reconstruction and generation quality. This objective encourages alignment with semantic features from the frozen VFM and effectively balances the contributions of low- and high-level features to the contextual tokens—thereby preserving semantic fidelity.
 \begin{wraptable}{r}{.62\textwidth}
    \caption{Ablation study on {\ours}'s components.}
	\label{tbl:ablation_queries}
	\centering
    \vspace{-.5em}
    \tablestyle{1.8pt}{1.05}
	\begin{tabular}{l|ccc|c|cc|c}
		\toprule
    \multirow{2}{*}{Setup} & \multicolumn{3}{c|}{Image Recon.} & \multicolumn{1}{c|}{Usage} & \multicolumn{2}{c|}{AR Gen.} & L.P. \\
    & \#Tok. & rFID$\downarrow$ & rIS$\uparrow$ & $\mathcal{Q}_{C}\uparrow$ & gFID$\downarrow$ & gIS$\uparrow$ & (\%) \\
    \midrule
    VQGAN & \multirow{2}{*}{576} &  0.95 & 197.3  & {99.7\%} & 3.71 & 228.3 & 23.1 \\
    + {Frozen VFM} &  & 0.99 & 206.3 &  100\% & 3.69 & 267.5 & 56.4 \\
    \midrule
    + {Region Adapt.} & \multirow{3}{*}{256} & 1.20 & 199.0 & \multirow{3}{*}{100\%} & 3.98 & 241.6 & 41.5 \\
    + {Multi-level Feat.} &  & 0.92 & 199.5 & & {3.71} & {251.1} & {22.7} \\
    + {Reconstruct Feat.} &  & \textbf{0.89} & \textbf{215.4} &  & \textbf{3.42} & \textbf{277.3} & \textbf{69.4}\\
    \midrule
    - {Frozen VFM} & 256 & 0.95 & 196.3 & 100\% & 3.73 & 248.7 & 59.1 \\
    \bottomrule
	\end{tabular}
    \vspace{-2em}
\end{wraptable}

With these three key components---(1) deformable attention for region-adaptive tokenization to reduce redundancy, (2) multi-level features for enhanced reconstruction, and (3) feature reconstruction loss for semantic alignment---VFMTok produces compact, semantically rich, and efficient tokens. Using only 256 tokens, VFMTok outperforms its VQGAN baseline with 576 tokens in reconstruction quality, generative performance, and semantic representation. Supplemental ablations are discussed in the Appendix.

\noindent\textbf{Convergence and efficiency analysis.}  
Beyond above analysis, we experiment VFMTok with a randomly initialized encoder instead of a pre-trained VFM with other components remaining unchanged. As shown in Tab.~\ref{tbl:ablation_queries} (last row), its reconstruction quality dropped to the level of VQGAN. Meanwhile, both its semantic representation capability and generation performance also decreased. This indicates a frozen VFM benefits tokenizer training as it provides a latent space advantageous for image reconstruction and generation. Besides, those semantic-rich, structured latent tokens accelerate AR model training convergence. As evidenced in Fig.~\ref{fig:pilot2}(b), VFMTok enables AR models to achieve a \textbf{3$\times$} speedup in convergence compared to VQGAN. Moreover, an AR model's generation time is quadratically proportional to the number of tokens. At the same resolution, VFMTok uses approximately half the tokens for image representation compared to counterparts like DINOv2-VQGAN and CLIP-VQGAN. Consequently, VFMTok achieves a \textbf{4$\times$} generation speedup over these counterparts depicted in Tab.~\ref{tbl:prelimiary_exp}. This acceleration can be further enhanced with CFG-free generation.

%% file: sections/5_conclusion.tex
\section{Conclusion}

In this work, we have demonstrated that frozen pre-trained vision foundation models (VFMs)---ranging from self-supervised to language-supervised-- are sufficient for high-quality image reconstruction and generation. To fully exploit their potential while addressing the redundancy inherent in 2D feature grids, we introduce \ours, a novel image tokenizer that incorporates region-adaptive tokenization to enhance token efficiency.
By reducing feature redundancy, integrating multi-level feature representations, and introducing a semantic-preserving feature reconstruction objective, \ours yields a compact and semantically rich latent space. This facilitates high-quality image reconstruction and generation, accelerates convergence in autoregressive (AR) models, and enables efficient, high-fidelity, classifier-free (CFG-free) image synthesis---without the need for additional heuristics. Furthermore, the reduced number of tokens significantly lowers the computational cost of AR inference, making the approach both scalable and effective.
Looking forward, the rich semantic structure of the learned latent space offers exciting potential for extending this work toward unified visual generation and understanding.

\section{Acknowledgments}

This work has been supported by the National Key R\&D Program of China (Grant No. 2022YFB3608300), Hong Kong Research Grant Council - Early Career Scheme (Grant No. 27209621), General Research Fund Scheme (Grant No. 17202422, 17212923, 17215025) Theme-based Research (Grant No. T45-701/22-R) and Shenzhen Science and Technology Innovation Commission (SGDX20220530111405040). Part of the described research work is conducted in the JC STEM Lab of Robotics for Soft Materials funded by The Hong Kong Jockey Club Charities Trust. We are deeply grateful to Lufan Ma for the contribution in polishing up this paper.
We would also appreciate Tong Yang for providing the DINO discriminator script.

\section{Author Contribution Statement}

X.Q. proposed the initial concept of region-adaptive quantization. Based on this, A.Z. built the VFMTok, conducted the overall experiments, and led the writing of the initial draft. X.W., X.Q., and C.M. were deeply involved in the project progress and manuscript writing. X(iangyu).Z., X(uanyang).Z., G.Y., and T.W., provided sufficient computational resources. X(uanyang).Z. joint discussion where suggested ablation studies along with T.W., and discussed the writing of the draft. All authors contributed critical feedback, shaping the research, analysis, and the final manuscript.

%% file: sections/6_appendix.tex
\section*{Appendix}
\vspace{-.3cm}
\appendix
This section first illustrates the implementation of VFMTok and AR generation models, covering their learning rates, optimization approaches, and training demands. Subsequently, we present more ablation studies on VFMTok, investigating the effects of different architectural designs on image reconstruction and generation. Besides, all VFMToks in this study utilize a frozen, pre-trained DINOv2-L~\cite{dinov2reg} as their encoder. Next, we show AR image generation with the other VFMs. Subsequently, we answer the question of what makes VFMs a good visual tokenizer. Finally, we demonstrate additional visualization samples of class-to-image generation using VFMTok.
\vspace{-.4cm}
\section{VFMTok Implementation}
\noindent\textbf{Tokenizer training.} VFMTok is trained on the ImageNet~\cite{imagenet} training set at $336\times336$ resolution with random crop augmentation. All models share identical training settings: a constant learning rate of ${10}^{-4}$, AdamW optimizer~\cite{adamw} (${\beta}_1 = 0.9, {\beta}_2 = 0.95$, weight decay = 0.05), a batch size of 256, and 50 training epochs. For training losses, the commitment loss weight is 0.25 and the adversarial loss weight is 0.5, with the adversarial loss activated after 20,000 iterations. Besides, VFMTok requires 1.5 days of training on 16 Nvidia H800 GPUs.

\noindent\textbf{AR model optimization.} The AR model training configuration aligns with LlamaGen's~\cite{llamagen}, except our training resolution is $336\times336$ and the duration depends on model parameters. Other key settings include: a base learning rate of ${10}^{-4}$ per 256 batch size; AdamW optimizer~\cite{adamw} (${\beta}_{1} = 0.9$, $\beta_{2} = 0.95$, weight decay = 0.05, gradient clipping of 1.0); a dropout rate of 0.1 for input token embeddings, attention, and FFN modules; and a 0.1 class condition embedding dropout for classifier-free guidance. Besides, A VFMTok-L requires 19.4 hours of training on 8 NVIDIA H800 GPUs.

\vspace{-.4cm}
\section{Supplemental Ablation Study}

\vspace{-.3cm}
In this subsection, we conduct more ablation studies on the design of our approach, including the AR image generation with resolution of $256\times256$, the effect of single-level \textit{v.s.} multi-level features, the effect of shared ViT \textit{v.s.} unshared ViT, the impact of different components, and the number of tokens to represent an image. Additionally, the effect of different VFMs on image reconstruction and generation is also presented.
\vspace{-.4cm}
\subsection{AR Generation with Resolution of 256×256}
\vspace{-.2cm}
In the main paper, given that the input resolution of the vision foundation model is $336\times336$, we adjust the resolution of reconstructed and generated images to $336\times336$ by default, thus avoiding changing the number of tokens for image representation. Following the common practices~\cite{llamagen, var}, we also train the image tokenizer and the AR generation model with the resolution of $256\times256$, respectively. We first initialize and train VFMTok tokenizer for 50 epochs, then integrate it with AR generative models. Considering computational costs, models with fewer parameters like VFMTok-B and VFMTok-L are trained for 300 epochs, while larger AR models for 50 epochs. This setting aligns with LlamaGen~\cite{llamagen} for $256\times256$ image generation. Furthermore, to ensure a fair comparison with the advanced autoregressive generation framework, RAR~\cite{rar}, we also incorporated VFMTok into RAR~\cite{rar}, maintaining the same setup during the training phase. As shown in \cref{tbl:low_resolution}, \ours not only achieves a decent reconstruction performance but also improves the generation quality compared to its counterparts~\cite{llamagen, rar} by a large margin. It is worth noting that \ours also accelerates the convergence speed during AR model training and significantly improves synthesis quality.


\begin{table*}[ht]
\vspace{-0.3cm}
	\caption{\ours performs {image reconstruction} and {AR generation} with the size of $256\times256$.}
	\label{tbl:low_resolution}
	\centering
    \vspace{-0.2cm}
    \setlength{\tabcolsep}{2.5pt}
	\begin{tabular}{l|c|cc|cc|c|c|cc}
		\toprule
            \multirow{2}{*}{Approach} & \multicolumn{3}{c|}{$\textit{Image recon.}$} &\multicolumn{2}{c|}{\textit{code usage}$\uparrow$} & \multicolumn{4}{c}{$\textit{AR gen.}$} \\
            \cmidrule{2-10}
            & \#Toks & rFID$\downarrow$ & rIS$\uparrow$ & $\mathcal{Q}_{C}$ & $\mathcal{Q}_{P}$ & \#$\text{Epochs}$ & Para. & gFID$\downarrow$ & gIS$\uparrow$   \\
            \hline
            LlamaGen-B  & \multirow{5}{*}{256} & \multirow{5}{*}{2.22} &\multirow{5}{*}{169.8} & \multirow{5}{*}{--} & \multirow{5}{*}{95.2\%} & \multirow{2}{*}{300} & 111M & 5.46 & 193.6  \\
            LlamaGen-L & &  &  &  &  & & 343M  & 3.81 & 248.3  \\
            \cline{7-7}
            LlamaGen-XL & &  &  &  &  & \multirow{3}{*}{50} & 775M  & 3.39 & 227.1  \\
            LlamaGen-XXL & &  &  &  &  & & 1.4B  & 3.09 & 253.6  \\
            LlamaGen-3B & & & & & & & 3.1B & 3.06 & 279.7  \\
            \midrule
            RAR-L~\cite{rar} & \multirow{3}{*}{256} & \multirow{3}{*}{2.12} & \multirow{3}{*}{171.4} & \multirow{3}{*}{--} & \multirow{3}{*}{100.0\%}  & \multirow{3}{*}{400} & 461M & 1.70 & 299.5  \\
            RAR-XL~\cite{rar} & & & & & & & 955M & 1.50 & 306.9 \\
            RAR-XXL~\cite{rar} & & & & & & & 1.5B & 1.48 & \textbf{326.0} \\
            \midrule
            \ours-B  & \multirow{7}{*}{256} & \multirow{7}{*}{1.02} & \multirow{7}{*}{213.2}  &\multirow{7}{*}{100.0\%}  & \multirow{7}{*}{--} &  \multirow{2}{*}{100} & 111M & {3.95} & {248.4} \\
            \ours-L & & &  &  & & & 343M  & {3.02} & {271.6} \\
            \cline{7-7}
            \ours-B & & &  &  & & \multirow{2}{*}{300} & 111M  & {3.61} & {247.6} \\
            \ours-L & & &  &  & & & 343M  & {2.79} & {276.0} \\
            \cline{7-7}
            \ours-XL & & &  &  & & \multirow{3}{*}{50} & 775M  & {2.79} & {277.1} \\
            \ours-XXL & & &  &  & & & 1.4B & {2.62} & {279.7} \\
            \ours-2B & & & & & & & 2B & {2.64} & {284.0}\\
            \midrule
            VFMTok(RAR-L) & \multirow{3}{*}{256} & \multirow{3}{*}{0.88} & \multirow{3}{*}{216.2} & \multirow{3}{*}{--} & \multirow{3}{*}{100.0\%}  & \multirow{3}{*}{400} & 461M & 1.47 & {316.2}  \\
            VFMTok(RAR-XL) & & & & & & & 955M & \underline{1.38} & 303.3 \\
            VFMTok(RAR-XXL) & & & & & & & 1.5B & \textbf{1.30} & 300.0 \\
		\bottomrule
	\end{tabular}
    \vspace{-0.4cm}
\end{table*}

\vspace{-.2cm}
\subsection{Effect of Single-level \textit{v.s.} Multi-level Features} 

In this paragraph, we investigate the separate impacts of each single-level or multi-level features on image reconstruction and generation. Specifically, we begin by conducting an ablation study to evaluate the impact of each single-layer feature on image reconstruction and generation. Subsequently, we cumulatively add each feature layer to observe the combined effect.

\begin{table}[htbp]
\centering
\begin{minipage}{0.49\textwidth}
\centering
\caption{Performance of each single-level feature. $F_i$ represents the indexed feature level.}
\label{tbl:each-single-level}
\vspace{3pt}
\tablestyle{1.5pt}{1.1}
\begin{tabular}{l|c|cc|c|cc}
\toprule
\multirow{2}{*}{$F_i$}  & \multicolumn{4}{c|}{$\textit{Image recon.}$} & \multicolumn{2}{c}{$\textit{AR gen.}$} \\
\cmidrule{2-5}\cmidrule{6-7}
&  \#Toks & rFID$\downarrow$ & rIS$\uparrow$ & $\mathcal{Q}_{C}$ & gFID$\downarrow$ & gIS$\uparrow$  \\
\midrule
$F_1$ & \multirow{4}{*}{256} & 1.04 & 186.4 & \multirow{4}{*}{100.0\%} & 3.84 & 257.9 \\
$F_2$ &   & 0.95 & 200.6 &  & 3.79	& 272.7 \\
$F_3$ &   & 1.03 & 208.5 & & 3.69 & 274.9 \\
$F_4$ &   &1.23  & 214.8 &   & 3.64 & 277.7  \\
\bottomrule
\end{tabular}
\end{minipage}
\hfill
\begin{minipage}{0.49\textwidth}
\centering
\caption{Performance of cumulatively added features. $F_i$ represents the indexed feature level.}
\label{tbl:cumulative-features}
\vspace{3pt}
\tablestyle{1.5pt}{1.1}
\begin{tabular}{l|c|cc|c|cc}
\toprule
\multirow{2}{*}{$F_i$}  & \multicolumn{4}{c|}{$\textit{Image recon.}$} & \multicolumn{2}{c}{$\textit{AR gen.}$} \\
\cmidrule{2-5}\cmidrule{6-7}
&  \#Toks & rFID$\downarrow$ & rIS$\uparrow$ & $\mathcal{Q}_{C}$ & gFID$\downarrow$ & gIS$\uparrow$  \\
\midrule
$F_1$ & \multirow{4}{*}{256} & 1.04 & 186.4 & \multirow{4}{*}{100.0\%} & 3.84 & 257.9 \\
$+F_2$ &   & 0.94 & 205.0 &  & 3.69	& 274.4 \\
$+F_3$ &   & 0.94 & 210.8 & & 3.27 & 272.5 \\
$+F_4$ &   & 0.89  & 215.4 &  & 3.09 & 274.2  \\
\bottomrule
\end{tabular}
\end{minipage}
\end{table}

\noindent\textbf{Setup.} We initialize and train each tokenizer for 50 epochs. Once the tokenizer is optimized, it is integrated with an AR generation model, \ours-L, for a total of 100 training epochs. During evaluation, the image reconstruction and generation performance are reported.

\noindent\textbf{Observation.} The results, as presented in the Table~\ref{tbl:each-single-level} and Table~\ref{tbl:cumulative-features}, show that while a single feature layer offers no significant advantage, the quality of both image reconstruction and generation markedly improves as more layers are incorporated.

\vspace{-.3cm}
\subsection{Effect of Shared ViT \textit{v.s.} Unshared ViT}

\begin{wraptable}{r}{0.6\textwidth}
\vspace{-0.45cm}
	\caption{Effect of shared ViT \textit{v.s.} unshared ViT.}
	\label{tbl:shared-vs-unshared}
	\centering
    \vspace{-0.23cm}
    \tablestyle{1.2pt}{1.1}
	\begin{tabular}{l|c|cc|c|c|cc|c}
		\toprule
           \multirow{2}{*}{Type}  & \multicolumn{3}{c|}{$\textit{Image recon.}$} &\multirow{2}{*}{$\mathcal{Q}_{C}$} & \multicolumn{3}{c|}{$\textit{AR gen.}$} & \multirow{2}{*}{L.P.} \\
            \cmidrule{2-4}\cmidrule{6-8}
            &  \#Toks & rFID$\downarrow$ & rIS$\uparrow$ &   & \#${E}$ & gFID$\downarrow$ & gIS$\uparrow$ &  \\
            \midrule
    unshared ViT  & \multirow{2}{*}{256} & {0.91} & {214.9} & {100.0\%} & 50 & 3.52 & {277.4} & {61.1}  \\
    \cline{1-1}\cline{3-9}
    shared ViT &  & {0.89} & {215.4} & {100.0\%} & 50 & {3.42} & 277.3 & {69.4} \\
		\bottomrule
	\end{tabular}
\vspace{-.4cm}
\end{wraptable}

In this work, VFMTok utilizes a shared ViT to generate latent features for pixel rendering and high-level VFM feature (specifically, from the last layer) reconstruction, respectively. However, it is uncertain if the sharing parameter is optimal. To this end, we experimented with another unshared ViT of the same architecture to generate the high-level VFM feature. Following the training setup, we train the tokenizer and AR model -- VFMTok-L for 50 epochs. As shown in Tab.~\ref{tbl:shared-vs-unshared}, the shared ViT ensures a better image reconstruction and synthesis quality with enriched semantics. Therefore, we utilize shared ViT in the VFMTok design by default.

\vspace{-.3cm}
\subsection{Effect of Different Components}

In \ours, we introduce three key features: a frozen foundation (\eg, DINOv2~\cite{dinov2reg}) model as the encoder, multi-level plain feature interaction, and reconstructing the pre-trained feature, to achieve a notable performance in image reconstruction and generation. To achieve this, we incorporate several learnable modules into \ours. Namely, the mask tokens ${M}_\text{I}$ and $\textbf{M}_{f}$ for image and its pre-trained feature reconstruction, the deformable attention layer adopted in the deformable transformer, and a set of learnable register tokens to address the potential artifacts in the latent feature. Thus, some questions naturally emerge: 1) Whether the mask token could be shared between ${M}_\text{I}$ and  $\textbf{M}_{f}$; 2) Could the deformable attention layer be replaced with the vanilla cross-attention layer; 3) Can the register tokens be discarded from \ours? 

\noindent\textbf{Setup.} To answer the above questions, we conducted 3 different experiments: 1) ${M}_\text{I}$ and $\textbf{M}_{f}$ share the same mask token $\text{M}_\text{share}$; 2) Replacing deformable attention with cross-attention in deformable transformer. To address the memory, computational demands, and fairness considerations of cross-attention, we first concatenate multi-level features from a VFM along the channel dimension, then apply a single MLP for dimensionality reduction before these features are fed to the cross-attention transformer for interaction with queries. Besides, each query interacts with VFM features within a $16\times16$ window to simulate region-adaptive behavior; 3) Removing the register tokens from \ours. Additionally, linear probing is carried out on the $\verb+[CLS]+$ token to estimate the semantic representation capability of \ours.

\noindent\textbf{Observation.}  
As shown in Tab.~\ref{tbl:part_annalysis}, unshared mask tokens seldom affect image reconstruction or generation quality but significantly degrade {\ours}'s overall semantic representation. This indicates that image reconstruction requires a certain amount of semantic information, but this semantic information is not as strong as that requested for VFM's feature reconstruction. Besides, Tab.~\ref{tbl:part_annalysis} also reveals that shared mask tokens can enhance the semantic representation of \ours, potentially benefiting downstream comprehension tasks. Hence, we use shared mask tokens by default in \ours. Introducing cross-attention for learnable queries and multi-level feature interaction shows an evident decline in image reconstruction, generation, and semantic level. Compared to deformable attention that focuses on local regions, using cross-attention may introduce redundant information, this redundancy weakens the overall semantic representation by complicating quantized visual token distribution, thereby hindering image generation. Additionally, a slight decrease in image reconstruction, generation, as well as the semantic representation is also observed when register tokens $\textbf{Tok}_\text{reg}$ are removed from \ours. This suggests that introducing register tokens into VFMTok is reasonable since they can eliminate some artifacts in the features.

\begin{table*}[htbp]
\vspace{-0.3cm}
	\caption{Impact study on mask embeddings, attention, and registered tokens.}
	\label{tbl:part_annalysis}
    \vspace{-.2cm}
	\centering
    \setlength{\tabcolsep}{5.0pt}
	\begin{tabular}{cccc|c|cc|cc|c}
		\toprule
           \multirow{2}{*}{$\text{M}_\text{shared}$}  & \multicolumn{2}{c}{Attention} & \multirow{2}{*}{$\textbf{Tok}_\text{reg}$} & \multicolumn{3}{c|}{$\textit{Image Recon.}$} & \multicolumn{2}{c|}{$\textit{AR gen.}$} & \multirow{2}{*}{L.P.} \\
            \cmidrule{2-3} \cmidrule{5-9}
            & Cross-Attn & Deform-Attn & & \#Toks & rFID$\downarrow$ & rIS$\uparrow$ & gFID$\downarrow$ & gIS$\uparrow$ & ~  \\
    \midrule
  ${\checkmark}$  & ${\checkmark}$ & & ${\checkmark}$ & \multirow{4}{*}{256} & 1.00 & 211.5 & 3.89 & 271.5 & 34.1 \\
 \checkmark   &  & \checkmark & &  & 0.91 & 215.8 & 3.45 & 276.5  & 68.3 \\
     & & \checkmark & \checkmark &  & 0.89 & {216.1} & 3.50 & 276.0 & 64.7 \\
   \checkmark & & \checkmark & \checkmark &  & {0.89} & {215.4} & {3.42} & {277.3} & {69.4} \\
		\bottomrule
	\end{tabular}
    \vspace{-.6cm}
\end{table*}

\subsection{Effect of the Number of Tokens to Represent an Image}

In this work, we utilize a set of region-adaptive tokens to represent an image. The number of these tokens is empirically fixed at 256 throughout our experiments. However, the impact of tokens' cardinality on image reconstruction and generation quality remains unexplored.

\begin{wraptable}{r}{0.37\textwidth}
        \vspace{-0.4cm}
	\caption{The effect of the number of tokens on image reconstructions and generation}
	\label{tbl:num_toks}
    \vspace{-.15cm}
	\centering
    \tablestyle{1.1pt}{1.05}
	\begin{tabular}{c|cc|cc}
		\toprule
           \multirow{2}{*}{\#Tokens}  & \multicolumn{2}{c|}{$\textit{Image recon.}$}  & \multicolumn{2}{c}{$\textit{AR gen.}$} \\
            \cmidrule{2-5}
            &   rFID$\downarrow$ & rIS$\uparrow$ & gFID$\downarrow$ & gIS$\uparrow$   \\
            \midrule
    36 & 2.61 & 175.4 & 3.93 & 222.4   \\
    64 & 2.09 & 188.3 & 3.59 & 250.5   \\
    100 & 1.44 & 204.4 & 3.54 & 270.8 \\
    121 & 1.36 & 202.5 & 3.59 & 267.1 \\
    {144} & {1.20} & {204.6} & {3.46} & {274.9} \\
    169 & 1.11 & 212.4 & 3.42 & 275.5 \\
    196 & 1.08 & 213.7 & 3.32 & 271.2 \\
    225 & 0.96 & 215.5 & 3.45 & 279.2 \\
    256 & 0.89 & 215.4 & 3.42 & 277.3 \\
    289 & 0.85 & 217.2 & 3.41 & 272.3 \\
    361 & 0.80 & 218.8 & 3.64 & 277.9 \\
    400 & 0.79 & 221.3 & 3.36 & 272.3  \\
    441 & 0.73 & 221.8 & 3.61 & 275.5 \\ 
    576 & 0.60 & 222.8 & 3.57 & 278.8 \\
		\bottomrule
	\end{tabular}
    \vspace{-.4cm}
\end{wraptable}

\noindent\textbf{Setup.} To investigate this, we designed a controlled setup to isolate the impact of query quantity while maintaining architectural consistency across experiments. Specifically, we parameterize the image tokenizer with varying numbers of learnable queries. Each tokenizer variant is trained on the ImageNet~\cite{imagenet} dataset for 50 epochs. Subsequently, the trained tokenizer is integrated into an AR image generation model, LlamaGen-L, and trained for 50 epochs. Both image reconstruction and generation quality are estimated on ImageNet~\cite{imagenet} dataset with FID and IS, respectively.

\noindent\textbf{Observation.} As show in Tab.~\ref{tbl:num_toks}, image reconstruction exhibits a positive correlation with the tokens' cardinality. As the number of tokens increases, metrics for estimating image reconstruction, namely rFID and rIS, both show gradual improvement. However, this trend does not hold for generation: as the number of tokens increases to higher values, there is no significant improvement in generation quality. Therefore, to balance generation quality with computational cost, and maintain fairness with vanilla counterparts~\cite{vqgan, vqgan-lc, llamagen}, we fix the number of tokens at 256 across our experiments. Actually, it's observed that \textbf{144} visual tokens suffice for representing images in ImageNet~\cite{imagenet}. This finding indicates VFMTok can further eliminate redundancy within image representations, yielding more compact and more effective image compression.


\subsection{Effect of Different VFMs}

Here, we explore employing different frozen VFMs~\cite{dinov2reg, siglip, siglip2} as the encoder of VFMTok, and incorporate them into different AR generative frameworks -- LlamaGen~\cite{llamagen} and RAR~\cite{rar}. As shown in Tab.~\ref{tbl:optimal_gen}, \ours implemented with SigLIP~\cite{siglip} also consistently yields decent image reconstruction and generation quality based on AR models of different scales. 
%
{Additionally, we also observe a potential correlation between the AR synthesis quality and different VFMs. VFM with stronger representation capabilities will achieve higher generation performance.}

\begin{table*}[ht]
\vspace{-0.2cm}
	\caption{Class-conditional image generation with different VFMs.}
	\label{tbl:optimal_gen}
	\centering
    \tablestyle{1.pt}{1.05}
	\begin{tabular}{l|l|cccccccc|ccccc}
    \toprule
\multirow{2}{*}{Type} & \multirow{2}{*}{{Method}} & \multirow{2}{*}{\#Epoch} & \multirow{2}{*}{{\#Para.}} & \multirow{2}{*}{{\#Tok.}} & \multicolumn{5}{c|}{{Generation w/ CFG}} & \multicolumn{5}{c}{{Generation w/o CFG}} \\
 & &  & & & {gFID} & {sFID} & {gIS} & {Pre.} & {Rec.} & {gFID} & {sFID} & {gIS} & {Pre.} & {Rec.}  \\
  \midrule
\multirow{13}{*}{AR.}& VFMTok-B(SigLIP) & \multirow{2}{*}{300} & 111M & \multirow{4}{*}{256} & 3.53 & 5.76 & 254.4 & 0.85 &  0.51 & 3.75 & 5.80 & 156.3 & 0.79 & 0.58 \\ 
~ & VFMTok-L(SigLIP) &  & 343M & ~ & 2.61 & 5.54 & 272.1 & 0.84 & 0.56 & 2.11 & 5.65 &  214.0 & 0.81 & 0.61 \\ 
& VFMTok-XXL(SigLIP) & \multirow{2}{*}{200}  & 1.4B & ~ & 2.09 & 5.75 & 272.6 & 0.82 & 0.60 & 1.85 & 5.78 & 251.2 & 0.81 & 0.61 \\
& VFMTok-2B(SigLIP) & ~ & 2.2B  & ~ & 2.05 & 5.77 & 271.4 & 0.82 & 0.61 & 1.92 &  5.78 & 260.5 & 0.82 & 0.61 \\
\cmidrule{2-15}
& VFMTok-XL(DINOv2) & \multirow{3}{*}{300} & \multirow{3}{*}{775M} & \multirow{3}{*}{256} & 2.41 & 5.53 & 276.8 & 0.83 & 0.59 & 2.10 & 5.54 & 258.1 & 0.82 & 0.60 \\
& VFMTok-XL(SigLIP) & ~ & ~ & ~ & 2.42 & 5.70 & 275.1 & 0.83 & 0.59 & {1.99} & 5.75 & 241.9 & 0.81 & 0.61  \\
& VFMTok-XL(SigLIP2) &    &      & ~ & 2.20 & \textbf{5.38} & 272.1 & 0.83 & 0.59 & {1.93} & 5.43 & 253.3 & 0.81 & 0.60 \\
\cmidrule{2-15}
~ & RAR-L(SigLIP) & \multirow{3}{*}{400} & 461M & \multirow{3}{*}{256} & 1.46 & 6.18 & \textbf{312.9}  & 0.78 & 0.64 & 2.26  & 5.78 & 204.3 & 0.79 & 0.62 \\
~ & RAR-XL(SigLIP) &  & 955M & & 1.40 & 6.39 & 311.7 &  0.78 & 0.65 & 1.87 & 5.77  & 226.7 & 0.79 &0.63 \\
~ & RAR-XXL(SigLIP) &  & 1.5B & & \textbf{1.38} & 6.18 & 298.4 & 0.78 & 0.65 &  1.68 & 5.64 & 239.5 & 0.79 & 0.63 \\
\cmidrule{2-15}
~ & RAR-XL(SigLIP2) & \multirow{3}{*}{400} & 461M & \multirow{3}{*}{256} & 1.50 &  6.37 & 292.7 & 0.77 & 0.66 & 2.05 &  5.50 & 217.7 & 0.79 & 0.62  \\
~ & RAR-XL(SigLIP2) &  & 955M & & 1.46 & 6.02 & 288.4  & 0.78 & 0.65 & 1.72 &  \textbf{5.41} &  244.8 & 0.80 & 0.63 \\
~ & RAR-XXL(SigLIP2) &  & 955M & & 1.43 & 6.07 & 291.0 &  0.79  &  0.65 &  \textbf{1.65} &  5.44 & \textbf{266.5} & 0.81 & 0.63 \\
  \bottomrule
	\end{tabular}
\vspace{-.5cm}
\end{table*}

\subsection{What makes a VFM a good visual tokenizer?}

To answer this problem, we first leverage existing vision foundation models, which were supervised with different learning objectives, \textit{e.g.} masked image modeling in pixel or latent space (Pixel-MIM and Latent-MIM) and contrastive learning (C.L) -- to construct VFMTok for image reconstruction and generation, and subsequently provide more insights on this question.

\noindent\textbf{Setup.} The encoder of vanilla VQGAN~\cite{llamagen} is substituted with different vision foundation models (VFMs), which are supervised with distinct learning objectives. All of these tokenizers are trained on the ImageNet~\cite{imagenet} training set for 50 epochs. Subsequently, these tokenizers are incorporated into an AR generative model -- LlamaGen-L~\cite{llamagen} and trained for 100 epochs. Both image reconstruction and generation quality are evaluated on the ImageNet validation set with FID and IS, respectively. Additionally, the top-1 accuracy of linear probing is also reported.

\noindent\textbf{Observation} As shown in the Table.~\ref{tbl:mim_gen}, Pixel-MIM (MAE~\cite{mae}) aligns best to reconstruction objectives, thus, VQGAN(MAE) achieves the optimal image reconstruction quality. However, it does not provide a feature space that is friendly for generation tasks. Its semantics is also inferior to the other VFMs. Contrastive learning-only models, VQGAN(CLIP) and VQGAN(SigLIP) achieve the best semantics (top-1 accuracy in linear probing), but modest image reconstruction and generation performance. Without a mask image modeling objective, they lack sufficient capability to model the local structures. Latent-MIM and contrastive learning co-optimized approaches, VQGAN(SigLIP2), and VQGAN(DINOv2) achieve the best overall performance.

\begin{table*}[ht]
	\caption{Image reconstruction and generation with VFM of a distinct learning objective.}
	\label{tbl:mim_gen}
	\centering
    \tablestyle{3.pt}{1.05}
	\begin{tabular}{l|ccc|ccccccc}
    \toprule
    Tokenizer & Pixel-MIM & Latent-MIM & C.L. & \#tok. & rFID$\downarrow$ & rIS$\uparrow$ & gFID $\downarrow$ & gIS$\uparrow$ & L.P.(\%)$\uparrow$\\ 
    \midrule
    VQGAN(CLIP~\cite{clip}) & \ding{55} & \ding{55} & \checkmark & \multirow{5}{*}{576} & 1.47 & 182.0 & 3.45 &221.2 &	59.5 \\
    VQGAN(SigLIP~\cite{siglip}) & \ding{55} & \ding{55} & \checkmark & & 1.26 & 190.8 & 3.50	& 246.1 &  60.3 \\
    VQGAN(SigLIP2~\cite{siglip2}) & \ding{55} & \checkmark & \checkmark & & 0.96 &  198.4 & 3.39 & 267.8 & 55.5 \\
    VQGAN(DINOv2~\cite{dinov2reg}) & \ding{55} & \checkmark & \checkmark &  & 0.99 & 206.3 & 3.34 & 268.6 & 56.4 \\
    VGQGAN(MAE~\cite{mae}) & \checkmark & \ding{55} & \ding{55} & & 0.67 & 207.6 &3.40&265.5	& 39.0 \\
  \bottomrule
	\end{tabular}
\vspace{-.3cm}
\end{table*}
\noindent\textbf{Discussion} The training objectives of vision foundation models(VFMs) affect the image reconstruction and generation of VFMTok. CLIP~\cite{clip} and SigLIP~\cite{siglip} apply contrastive learning to image-text pairs, whereas DINO~\cite{dino} applies a contrastive learning-like clustering objective to images. MAE~\cite{mae} performs mask image modeling on image patches in pixel space, and iBOT~\cite{ibot} predicts DINO-like cluster assignments of image patches in latent space. DINOv2~\cite{dinov2reg} basically is the combination of DINO~\cite{dino} and iBOT~\cite{ibot}, and SigLIP2~\cite{siglip2} is also essentially the integration of SigLIP~\cite{siglip} and iBOT~\cite{ibot} (termed TIPS loss in their paper), omitting other regularizing losses. We hereby provide a holistic comparison of the training objectives of different VFMs in Table~\ref{tbl:sup_vfm}. 

\begin{wraptable}{r}{0.35\textwidth}
	\caption{The category of learning objectives adopted in VFMs.}
	\label{tbl:sup_vfm}
    \vspace{-.15cm}
	\centering
    \tablestyle{1.1pt}{1.05}
	\begin{tabular}{l|ccc}
		\toprule
        VFM & P-MIM & L-MIM & C.L. \\
        \midrule
        CLP~\cite{clip} & \ding{55} & \ding{55} & \checkmark \\
        DINO~\cite{dino} & \ding{55} & \ding{55} & \checkmark  \\
        SigLIP~\cite{siglip} &  \ding{55} & \ding{55} & \checkmark \\
        MAE~\cite{mae} & \checkmark & \ding{55} & \ding{55} \\
        iBOT~\cite{ibot} & \ding{55} & \checkmark & \ding{55} \\
        SigLIP2~\cite{siglip2} & \ding{55} & \checkmark & \checkmark \\
        DINOv2~\cite{dinov2reg} & \ding{55} & \checkmark & \checkmark \\
		\bottomrule
	\end{tabular}
    \vspace{-.4cm}
\end{wraptable}
Notably, iBOT~\cite{ibot}, DINOv2~\cite{dinov2reg}, and SigLIP2~\cite{siglip2} are performing codebook learning resemble that in VQGAN on image patches, despite codes being soft instead of one-hot, and their codebook vectors being high-dimensional (\textit{e.g.}, 256). Following DINO~\cite{dino}, they learn a set of 8,192 or 65,536 prototypes, and require two augmented versions of the same patch (or image) to have identical prototype assignments (soft codes). They also predict the soft codes of masked patches, thus performing mask image modeling(Latent-MIM) in latent space. This formulation is termed self-distillation in DINO~\cite{dino}, and then extended to patches with Laent-MIM by iBOT~\cite{ibot}, which was subsequently inherited by DINOv2~\cite{dinov2reg} and SigLIP2~\cite{siglip2}.

The connection between Latent-MIM and VQGAN makes it natural to convert these VFMs to tokenizers and expect decent image reconstruction and generation performance. Additionally, the global (image-level) contrastive learning objectives of DINOv2~\cite{dinov2reg} and SigLIP2~\cite{siglip2} also promise better high-level semantics (better performance on understanding tasks).

To summarize, the conclusion can be drawn as: masked image modeling(MIM) objectives primarily help reconstruction and generation. MIM in pixel space helps reconstruction more, but is less beneficial for generation than MIM in latent space. Contrastive learning(C.L.) objectives are less helpful for reconstruction and generation, but are important for understanding abilities (\textit{e.g.}, top-1 accuracy on ImageNet~\cite{imagenet}). Best VFMs for visual tokenization are trained with both mask image modeling in latent space and contrastive objectives, namely DINOv2~\cite{dinov2reg} and SigLIP2~\cite{dinov2reg}.

\vspace{-0.3cm}
\section{Limitations}
Beyond the decent improvement achieved by VFMTok the optimal architectural design of VFMTok and its scalability on large-scale datasets are still under exploration. Currently, VFMTok's image reconstruction is not yet optimal, primarily due to limitations in its codebook design and model size. For fair comparison and maintaining a concise design, we adopt a vanilla codebook and keep VFMTok's size comparable to VQGAN's~\cite{llamagen}. While advanced codebook designs and increased model size could further enhance reconstruction quality. Beyond architecture design, exploring the scalability of VFMTok on large-scale datasets is also crucial for developing a superior tokenizer and advancing towards unified generation and understanding tasks. These aspects, however, outline promising directions for our future research.


\vspace{-0.3cm}
\section{Broader Impacts}

The advancements in image tokenizers and autoregressive (AR) image generation present significant broader impacts. Positively, these technologies can democratize content creation across art, design, and media, accelerate scientific research through enhanced data augmentation and visualization, and enable novel forms of personalized digital experiences. However, this transformative potential is accompanied by substantial ethical challenges. Key concerns include the proliferation of realistic misinformation (deepfakes), the potential for misuse in creating non-consensual or harmful content, the amplification of societal biases embedded in training data, and complex issues surrounding intellectual property and copyright. Furthermore, the computational resources required for training state-of-the-art models raise environmental considerations. Therefore, the responsible development and deployment of these powerful tools necessitate robust frameworks for ethical guidelines, bias detection and mitigation, provenance tracking, and public awareness to harness their benefits while minimizing societal risks.

\begin{figure}[thbp]
  \centering
  \includegraphics[width=\textwidth]{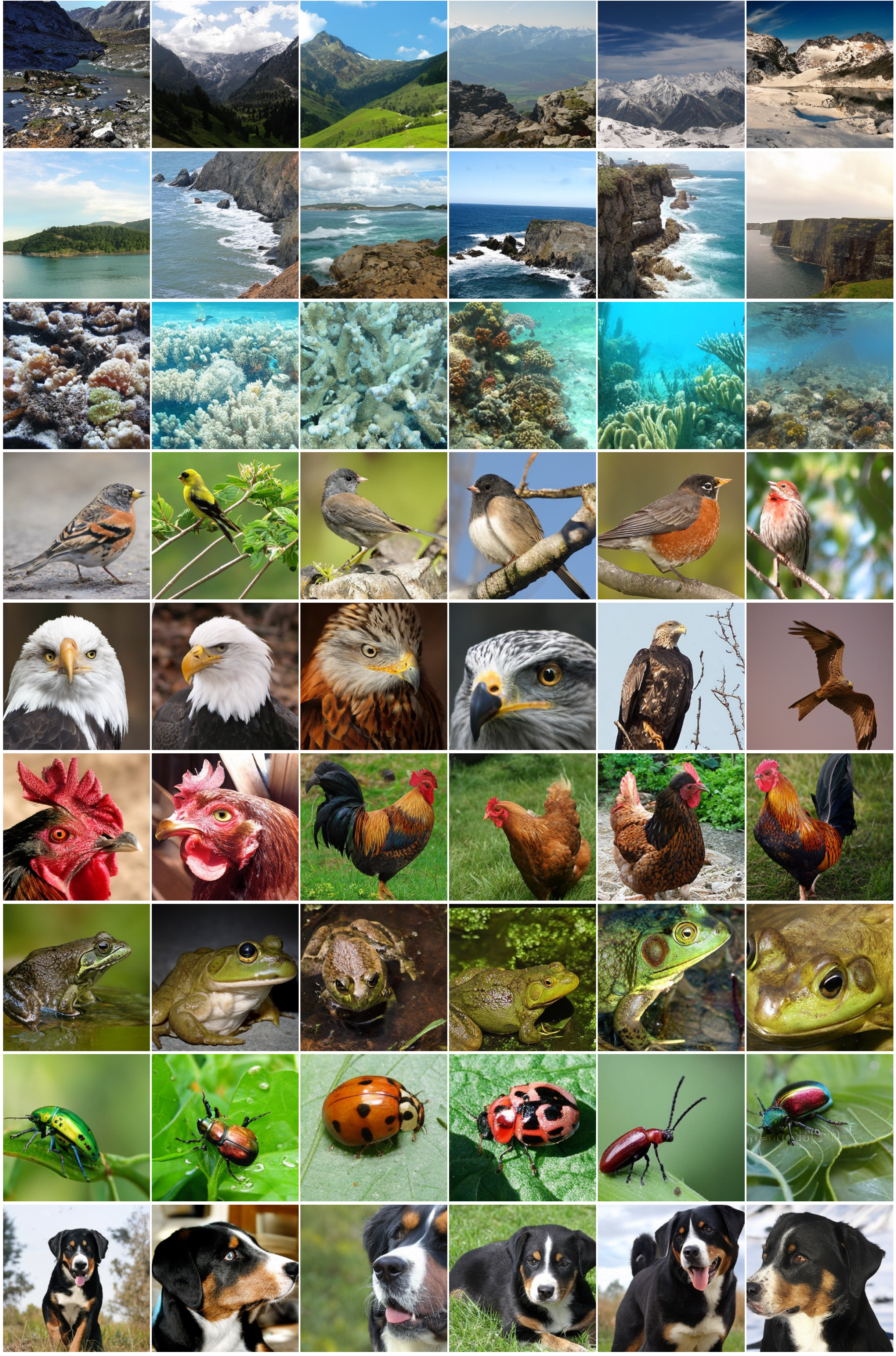}
  \vspace{-0.6cm}
  \caption{Class-conditional image generation with CFG.}
  \label{fig:supp_gen_cfg}
\end{figure}

\begin{figure}[thbp]
  \centering
  \includegraphics[width=\textwidth]{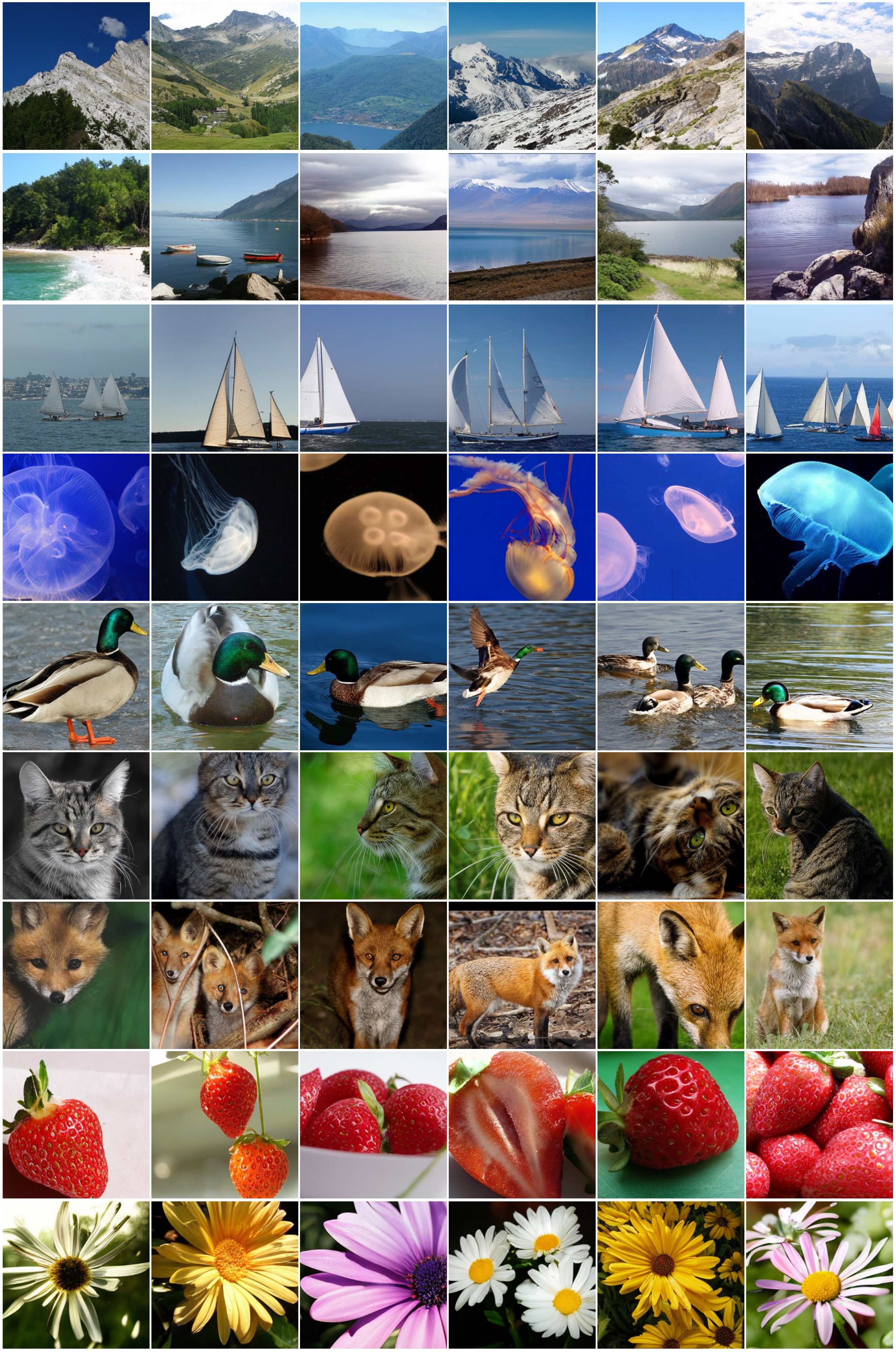}
  \vspace{-0.6cm}
  \caption{Class-conditional image generation without CFG.}
  \label{fig:supp_gen_wo_cfg}
\end{figure}